\documentclass[10pt,twocolumn,letterpaper]{article}

\usepackage[pagenumbers]{cvpr} 
\usepackage{multirow}
\definecolor{cvprblue}{rgb}{0.21,0.49,0.74}
\usepackage[pagebackref,breaklinks,colorlinks,allcolors=cvprblue]{hyperref}
\usepackage{graphicx}   
\usepackage{caption}    
\usepackage[accsupp]{axessibility} 
\usepackage{multirow}
\usepackage{wrapfig}
\usepackage{booktabs}
\usepackage{bbding}
\usepackage[table]{xcolor} 
\def\paperID{*****} 
\def\confName{CVPR}
\def\confYear{2026}

\title{Text-Phase Synergy Network with Dual Priors for Unsupervised Cross-Domain Image Retrieval}

\author{
Jing Yang$^{1,2}$ \quad
Hui Xue$^{1,2}\thanks{Co-corresponding authors}$ \quad
    Shipeng Zhu$^{1,2}$ \quad
    Pengfei Fang$^{1,2}$\footnotemark[1] \\
    $^{1}$School of Computer Science and Engineering, Southeast University, China \\
    $^{2}$Key Laboratory of New Generation Artificial Intelligence Technology and \\ Its Interdisciplinary Applications  (Southeast University), Ministry of Education, China \\
    {\tt\small \{yangjing.seu, hxue, shipengzhu, fangpengfei\}@seu.edu.cn}}

\begin{document}
\maketitle
\begin{abstract}
This paper studies unsupervised cross-domain image retrieval (UCDIR), which aims to retrieve images of the same category across different domains without relying on labeled data. Existing methods typically utilize pseudo-labels, derived from clustering algorithms, as supervisory signals for intra-domain representation learning and cross-domain feature alignment. However, these discrete pseudo-labels often fail to provide accurate and comprehensive semantic guidance. Moreover, the alignment process frequently overlooks the entanglement between domain-specific and semantic information, leading to semantic degradation in the learned representations and ultimately impairing retrieval performance. This paper addresses the limitations by proposing a \textit{Text-Phase Synergy Network with Dual Priors} (TPSNet). Specifically, we first employ CLIP to generate a set of class-specific prompts per domain, termed as domain prompt, serving as a \textit{text prior} that offers more precise semantic supervision. In parallel, we further introduce a \textit{phase prior}, represented by domain-invariant phase features, which is integrated into the original image representations to bridge the domain distribution gaps while preserving semantic integrity. Leveraging the synergy of these dual priors, TPSNet significantly outperforms state-of-the-art methods on UCDIR benchmarks. 
\end{abstract}    
\section{Introduction}
\label{sec:intro}

Cross-domain image retrieval aims to bridge the distribution gaps between heterogeneous image domains (e.g., real images and sketches) and enable accurate matching by extracting domain-invariant semantic features. It has broad applicability across various fields, including intelligent security~\cite{face}, medical image analysis~\cite{medical}, mobile product image search~\cite{mobile}, to name but a few. Most existing methods address this task by training a retrieval model using extensive annotated data~\cite{cdir1,cdir4,cdir5}. However, in real-world scenarios, acquiring large-scale, high-quality labeled data is often prohibitively expensive. To mitigate the cost of manual annotation, an emerging solution is to learn directly from unlabeled data, a task known as unsupervised cross-domain image retrieval (UCDIR).
\begin{figure}[t!]
    \centering
    \includegraphics[width=1\linewidth]{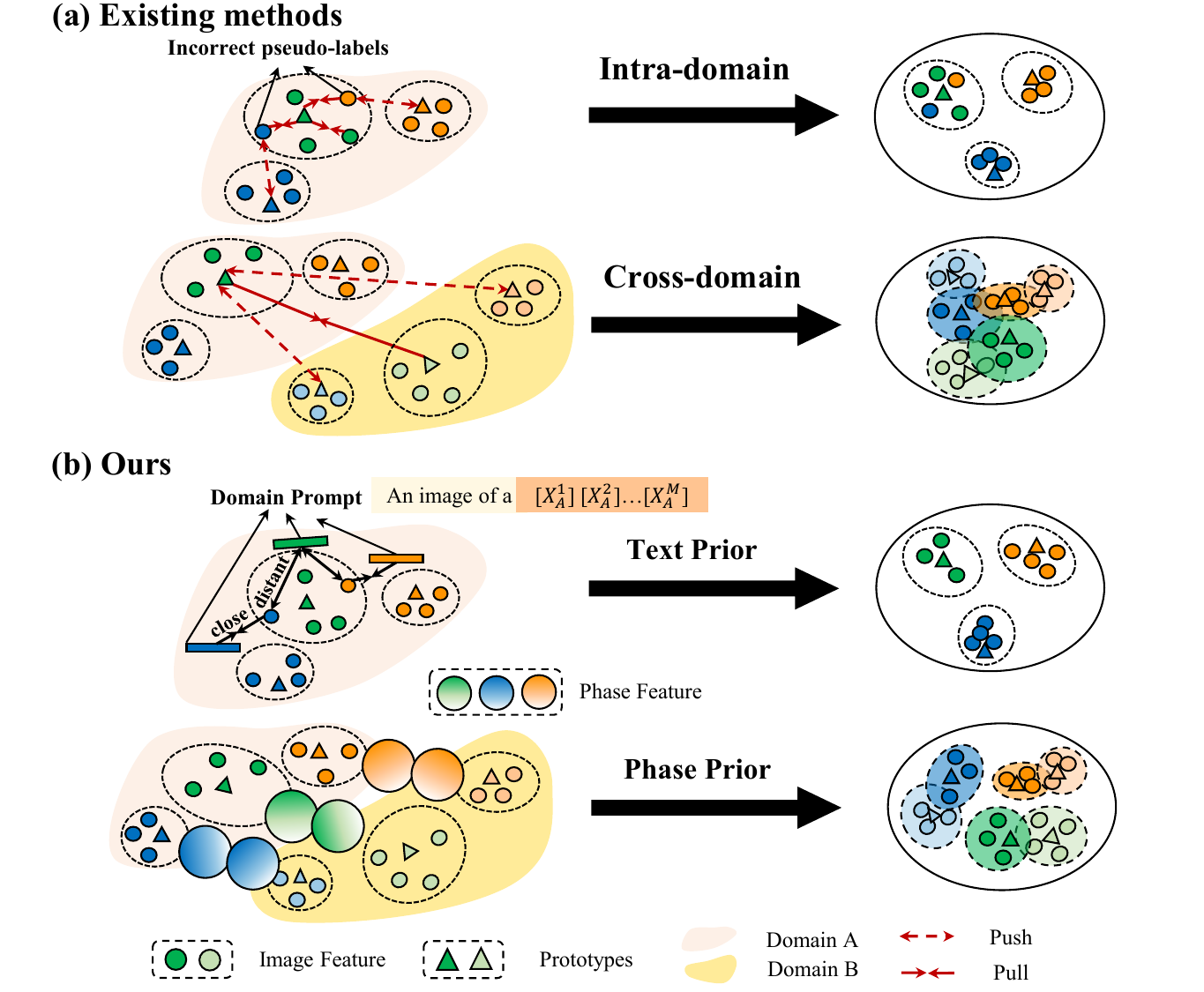}
    \vspace{-1em}
    \caption{Comparison between (a) existing methods and (b) our proposed TPSNet. Existing methods rely on inaccurate pseudo-labels for intra-domain and cross-domain learning, often causing semantic loss. In contrast, TPSNet leverages text and phase dual priors to extract domain-invariant semantic features.}
    \vspace{-1em}
    \label{figure1}
\end{figure}

In UCDIR, the absence of label information poses a significant challenge to learn domain-invariant semantic features for effective cross-domain retrieval. To address this issue, recent approaches commonly adopt a two-stage learning paradigm: intra-domain representation learning followed by cross-domain feature alignment. This is typically achieved by first applying a clustering algorithm to generate pseudo-labels, and then constructing class prototypes for each cluster to facilitate the learning process. However, it introduces two key issues, as illustrated in Figure~\ref{figure1}. In the intra-domain representation learning stage, contrastive learning is reliant on pseudo-labels to pull together images from the same category and push apart those from different categories, thereby capturing semantic structure in each domain of data. Nevertheless, the resulting discrete pseudo-labels often fail to provide accurate and comprehensive semantic supervision. Such weak supervision leads to improper pairwise constraints and less discriminative class prototypes, which in turn hinder prototype-based alignment and degrade the quality of learned feature representations. In the cross-domain feature alignment stage, techniques such as adversarial domain adaptation~\cite{semantic}, statistical distribution alignment~\cite{ucdir}, and cross-domain contrastive learning~\cite{protoot} are commonly used to reduce domain discrepancies from the perspective of the spatial domain. However, these alignment strategies may degrade semantic information due to the entangled nature of domain-specific and semantic features~\cite{DA2}, especially in cases where pseudo-labels and prototypes are inaccurate or unreliable.
The recent success of vision-language models (VLMs), such as CLIP~\cite{clip}, has demonstrated promising capabilities in learning cross-modal semantic representations through large-scale image-text alignment~\cite{vilt,VLR1}. These textual representations provide richer semantic priors than discrete pseudo-labels, and can guide visual feature learning via multi-modal interaction~\cite{Styleclip,Denseclip,prompt}. Meanwhile, another line of research seeks to mitigate domain discrepancies by manipulating frequency components in the frequency domain, rather than relying on explicit alignment in the spatial domain. For example, studies such as FDA~\cite{FDA} and FUDA~\cite{FUDA} obtain domain-invariant representations by replacing the low-frequency components of source images with those from target images, reducing inter-domain discrepancies while largely preserving semantic content. However, most existing approaches focus on manipulating high-frequency and low-frequency components. In contrast, the spectral decomposition perspective that separates amplitude and phase spectra remains underexplored, even though the phase spectrum is known to encode structural and semantic cues that are more robust to domain shifts~\cite{phase}.

Inspired by recent studies, we propose the \textit{Text-Phase Synergy Network with Dual Priors} (TPSNet) for UCDIR, with its key components illustrated in Figure~\ref{figure1}. We first introduce domain prompts, initialized as learnable embeddings for each cluster based on pseudo-labels generated via $K$-means clustering. The prompts are further tuned through an improved unsupervised CLIP-based contrastive learning scheme. It incorporates the semantic information encoded in the prompts to attain refined pseudo-labels. The resulting domain prompts then serve as a \textit{text prior}, providing stronger semantic supervision for subsequent representation learning. Recognizing that the phase spectrum in the frequency domain conveys semantic signals invariant across data domains~\cite{phase,dgucdir}, we additionally incorporate phase features as a \textit{phase prior} to bridge domain discrepancies while preserving essential semantic content, often degraded by traditional alignment strategies. By jointly leveraging these \textit{text-phase dual priors}, TPSNet effectively captures domain-invariant semantic features, thereby significantly enhancing performance on the UCDIR task. The contributions of this work can be summarized as follows:
\begin{itemize}
\item We introduce learnable domain prompts as a text prior to mitigate pseudo-label noise in UCDIR, enhancing semantic feature extraction via cross-modal alignment.
\item We further incorporate a phase prior to alleviate semantic degradation in cross-domain alignment, enabling robust and domain-invariant representations through dual priors.
\item Extensive experiments on the UCDIR datasets demonstrate the effectiveness of our method, which performs significantly better than the current state-of-the-arts and provides a new solution for the field of UCDIR.
\end{itemize}

\section{Related Work}
\label{sec:related work}

\paragraph{Cross-Domain Image Retrieval.}
Cross-domain image retrieval involves retrieving images where the query and gallery sets originate from different domains. Existing methods~\cite{cdir1,cdir2,cdir6} predominantly rely on supervised learning with labeled data, which limits scalability in real-world applications. To address this limitation, recent research has focused on the setting of unsupervised cross-domain image retrieval (UCDIR), which seeks to learn domain-invariant semantic representations from unlabeled data. Representative approaches such as DD~\cite{ucdir} and CODA~\cite{coda} employ $K$-means clustering to generate pseudo-labels. DD further utilizes these pseudo-labels for intra-domain contrastive learning and minimizes the Distance-of-Distance loss to align distributions, while CODA leverages pseudo-labels to initialize both intra-domain and cross-domain classifiers, enabling the separate learning of intra-domain representations and cross-domain feature alignment. However, these methods decompose the UCDIR task into two independent stages, intra-domain representation learning and cross-domain alignment, overlooking the intrinsic correlation between them. To address this, ProtoOT~\cite{protoot} and ShieldIR~\cite{shieldir} unify the two stages within an optimal transport framework, combining prototype alignment with contrastive learning. However, semantic degradation persists due to the entanglement of domain-specific and semantic features. SA-MoE~\cite{samoe} addresses this issue by emphasizing semantic modeling through semantically attentive fusion and contextual relevance, thereby preserving semantic integrity and underscoring the importance of semantics in UCDIR.

\paragraph{Unsupervised Domain Alignment Methods.}
Unsupervised domain alignment aims to bridge the distribution gaps between domains in the absence of labeled data. Within the UCDIR framework, effective domain alignment is critical for robust retrieval performance under domain discrepancies. To tackle this challenge, various alignment strategies have been proposed. Statistical alignment methods, such as MMD~\cite{MMD} and MDD~\cite{MDD}, explicitly reduce domain discrepancies by aligning statistical properties (e.g., means and covariances) across domains. In contrast, adversarial learning approaches, including DANN~\cite{DANN} and CDAN~\cite{CDAN}, utilize domain discriminators to encourage feature extractors to learn domain-invariant representations through adversarial training. More recently, Optimal Transport (OT) has been applied to align domain distributions by computing an optimal mapping that minimizes transport cost~\cite{OT1,OT2}, providing a foundation for subsequent cross-domain contrastive learning that further mitigates domain discrepancies. Diverging from conventional alignment techniques, another line of work~\cite{FDA,FUDA} uses frequency domain information to reduce distribution discrepancies while preserving essential semantic content, offering a novel perspective for cross-domain alignment.

\section{Method}

\label{sec:method}
\subsection{Overview}
We first formulate the problem of the UCDIR. Given two sets of unlabeled training images from different domains, denoted as $\mathcal{D}_A=\{x_i^A\}_{i=1}^{N_A}$ and $\mathcal{D}_B=\{x_j^B\}_{j=1}^{N_B}$, the goal of UCDIR is to learn a shared embedding space where semantically similar images from both domains are projected onto nearby representations. During testing, given a query image $x_i^A \in \mathcal{D}_A$ with class label $y_i$, the model is tasked with retrieving all semantically relevant images from $\mathcal{D}_B$ that share the same label $y_i$.

\begin{figure*}[t]
    \centering
    \includegraphics[width=\linewidth]{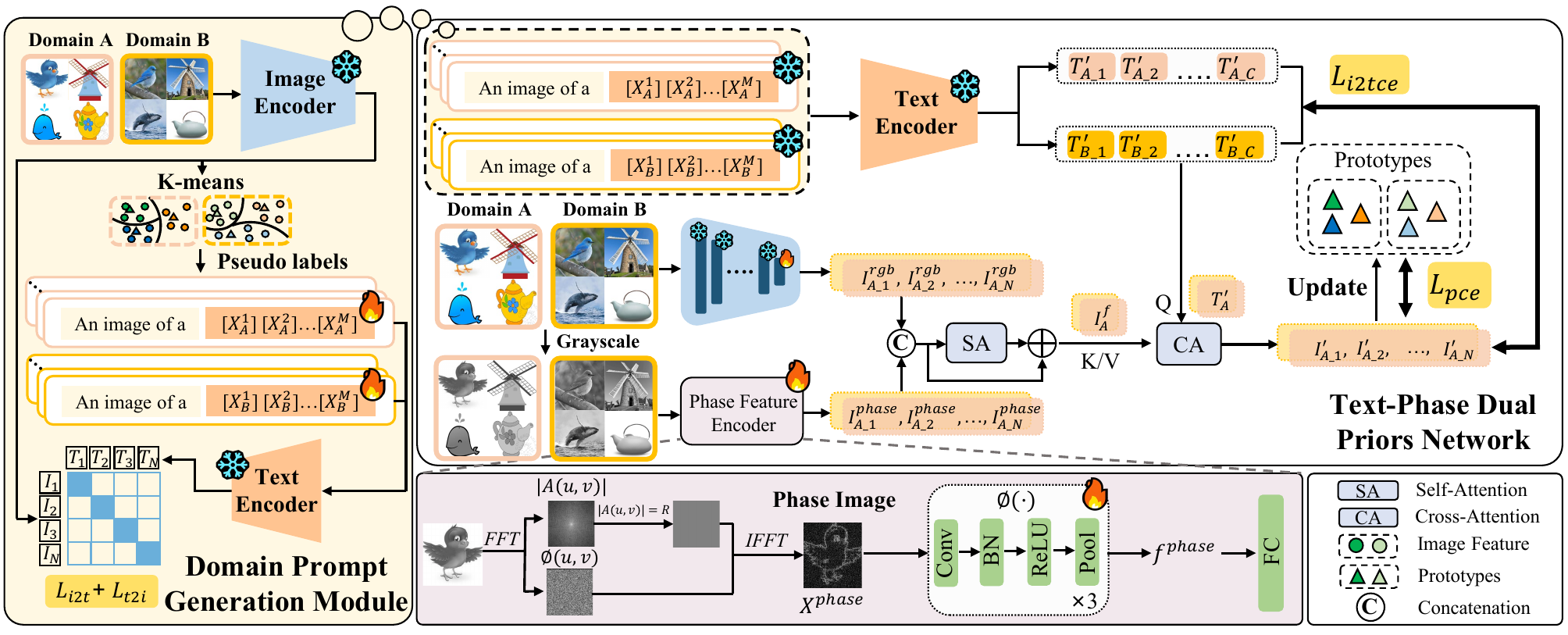}
    \caption{The pipeline of TPSNet. Left: domain prompt generation via the prompt learning paradigm. 
    Top-right: text-phase dual prior construction with contrastive learning for unsupervised cross-domain image retrieval. 
    Bottom: the detailed architecture of the proposed phase feature encoder.}
    \label{figure2}
    \vspace{-0.5em}
\end{figure*}


UCDIR is a challenging problem due to the stringent requirement of training a retrieval model in different data domains without relying on any annotations. To address this, we propose a novel approach named TPSNet, as illustrated in Figure~\ref{figure2}. TPSNet comprises two key components: one Domain Prompt Generation Module (DPG) and one Text-Phase Dual Priors Network (TPDP). In DPG, class-specific prompts are optimized per domain using CLIP-based contrastive learning. Having the learned domain prompts at hand, TPDP adopts a dual-path architecture: the text branch leverages the learned domain prompts to provide enriched semantic supervision, while the image branch integrates phase features to support cross-domain alignment and preserve semantic content. By leveraging the synergy of text-phase dual priors, TPSNet effectively facilitates the extraction of domain-invariant semantic visual representations.

\subsection{Domain Prompt Generation Module}
The overall diagram of the proposed DPG is presented on the left of Figure~\ref{figure2}. Following previous UCDIR methods, we first employ $K$-means clustering to generate discrete pseudo-labels for two domain of images. 
For each domain, we initialize $C$ learnable class-specific prompts corresponding to the clusters identified by pseudo-labels. Each prompt is formulated as a textual template, e.g., ``An image of a $[X]_A^1[X]_A^2...[X]_A^M$.'' for $\mathcal{D}_A$ and ``An image of a $[X]_B^1[X]_B^2...[X]_B^M$.'' for $\mathcal{D}_B$. Here, $[X]^i$ is a learnable token that is randomly initialized, and $M$ denotes the number of learnable tokens.



The initialized domain prompts and images are then fed into the pre-trained CLIP model as inputs, where a cross-modal contrastive learning strategy is employed to effectively update the domain prompts. Note that the parameters of both the image and text encoders are frozen, with only the token $[X]^i$ within the domain prompts is optimized. During this stage, the contrastive losses $\mathcal{L}_{i2t}^i$ and $\mathcal{L}_{t2i}^i$ are utilized, defined as follows:
\begin{equation}
\mathcal{L}_{i2t}^i=-\log \frac{\exp \left(s\left(I_{i}, T_{i}\right)/\tau\right)}{\sum_{j=1}^{N} \exp \left(s\left(I_{i}, T_{j}\right)/\tau\right)},
\label{eq:1}
\end{equation}
and
\begin{equation}
\mathcal{L}_{t2i}^i=-\frac{1}{\left|P\left(y_{i}\right)\right|} \sum_{p \in P\left(y_{i}\right)} \log \frac{\exp \left(s\left(I_{p}, T_{y_{i}}\right)/\tau\right)}{\sum_{j=1}^{N} \exp \left(s\left(I_{j}, T_{y_{i}}\right)/\tau\right)},
\label{eq:2}
\end{equation}
where $I_{i}$ and $T_{i}$ denote the re-paired image and text embeddings based on cosine similarity scores, which allows the model to partially correct inaccurate pseudo-labels through contrastive alignment. $s(I_{i}, T_{i})= \frac{I_i^{\top} T_i}{\|I_i\| \|T_i\|}$ and $\tau$ is the temperature. $N$ is the batch size, $P(y_{i})$ is the set of positive samples with respect to $T_{y_{i}}$, i.e., $P(y_{i})=p\in\{1 ... N|y_p=y_i\}$. The symbol $| \cdot |$ denotes the Cardinality of this set. Then the proposed DPG is optimized via the following loss: $\mathcal{L}_{prompt}=\mathcal{L}_{i2t}^i+\mathcal{L}_{t2i}^i$.


By minimizing $\mathcal{L}_{{prompt}}$, the incrementally optimized domain prompts are leveraged as supervisory signals to refine the potentially inaccurate discrete pseudo-labels generated by the clustering process in DPG. This optimization yields domain prompts that encapsulate semantic information from the text representations, which are subsequently incorporated into the TPDP module to guide the extraction of semantic representations.

\subsection{Text-Phase Dual Priors Network}


The overall architecture of TPDP is illustrated in the top-right part of Figure~\ref{figure2}. The proposed method adopts a dual-path design, consisting of a Text-Prior Semantic Feature Extraction (TPFE) module for the text path and a Phase-Prior Domain-Invariant Feature Extraction (PPFE) module for the image path. The following sections provide detailed descriptions of TPFE and PPFE, respectively.

\paragraph{Text-Prior Semantic Feature Extraction.}
In order to  extract domain-invariant semantic features, we design the TPFE module utilizing the domain prompts learned from DPG. These domain prompts are passed through a shared text encoder to generate text features. For $\mathcal{D}_A$, the resulting text features are denoted as $T_A^\prime=\{T_{A\_1}^\prime, T_{A\_2}^\prime, \dots, T_{A\_C}^\prime\}$, and for $\mathcal{D}_B$, as $T_B^\prime=\{T_{B\_1}^\prime, T_{B\_2}^\prime, \dots, T_{B\_C}^\prime\}$, where $C$ is the number of semantic categories. These text features, $T_A^\prime$ and $T_B^\prime$, will serve as text prior to guide the semantic representation learning in subsequent process.

\paragraph{Phase-Prior Domain-Invariant Feature Extraction.}

In the PPFE module, images from two domains $\mathcal{D}_A$ and $\mathcal{D}_B$ are first processed by an image encoder to obtain RGB features $I^{rgb}$. Although $I^{rgb}$ provides high-level semantics, it is inevitably influenced by domain-specific factors such as background and style, which limits its effectiveness for cross-domain retrieval. To further mitigate this issue, we further extract the corresponding phase features $I^{phase}$ from grayscale images using a dedicated Phase Feature Encoder, as described in the following paragraph, and fuse them with $I^{rgb}$:
\begin{equation}
\begin{aligned}
I^f &= \text{Mean}\Big( 
    \text{LayerNorm}\big( 
        \text{Concat}(I^{rgb}, I^{phase}) \\
    &\quad + \text{SelfAttention}\big( \text{Concat}(I^{rgb}, I^{phase}) \big) 
    \big) 
\Big),
\end{aligned}
\end{equation}
where Concat(·) and Mean(·) denote the concatenation and mean pooling, respectively. SelfAttention(·) and LayerNorm(·) refer to the self-attention and layer normalization. This operation yields the fused features $I^f$ for two domains, effectively integrating semantic content from RGB with enhanced structural consistency from phase to fuse domain-invariant information while retaining rich semantics for robust representations.

\paragraph{Phase Feature Encoder.} 
The architecture of the Phase Feature Encoder is illustrated at the bottom of Figure~\ref{figure2}. In this process, we apply the Fast Fourier Transform (FFT) to the grayscale images to obtain the frequency representation $F(u,v) = |A(u,v)|e^{j\phi(u, v)}$, where $|A(u,v)|$ and $\phi(u,v)$ denote the amplitude spectrum and phase spectrum, respectively. To emphasize domain-invariant information, we retain the phase spectrum while replacing the amplitude spectrum with a constant value $R$, yielding the modified frequency signal $F'(u,v) = R e^{j\phi(u,v)}$. The final phase image $X^{phase}$ is reconstructed by applying the Inverse Fast Fourier Transform (IFFT):
\begin{equation}
    X^{phase} = \mathcal{F}^{-1}\{F'(u,v)\},
\end{equation}
where $\mathcal{F}^{-1}$ denotes the IFFT operation. Then, we design a lightweight convolutional subnetwork to extract phase information $f^{phase}$ from the phase image $X^{phase}$:
\begin{equation}
    \phi(\cdot)=\text{AvgPool}_{2\times2}(\text{ReLU}(\text{BN}(\text{Conv}_{3\times3}(\cdot)))),
\end{equation}
\begin{equation}
    f^{phase} = \phi\left( X^{phase} \right).
\end{equation}

The $f^{phase}$ is then passed through a fully connected layer, resulting the final phase features as: 
\begin{equation}
    I^{phase} = \text{FC}\left( \text{Flatten}(f^{phase}) \right).
\end{equation}

\paragraph{Synergy of Text-Phase Dual Priors.}
To effectively incorporate dual priors from text and phase information into cross-domain retrieval, we adopt a cross-attention mechanism to enable their synergistic integration. For each domain, the resulting text features $T^\prime$ serve as query vectors and the fused visual features $I^f$ are used as key and value vectors. The cross-attention mechanism is adopted to fuse the two priors, as:
\begin{equation}
I^\prime = \text{CrossAttention}( T^\prime; I^f).
\end{equation}

This yields the final visual features $I^\prime_A$ and $I^\prime_B$ for the two domains, which are enriched with the synergy of text-phase dual priors and retain domain-invariant semantic information. After obtaining the final visual features for two domains, we perform prototype updates at each training epoch for each domain individually:
\begin{equation}
    \mathcal{P}_{y_{i}}\leftarrow m \mathcal{P}_{y_{i}}+(1-m)I_{i}^\prime,
\end{equation}
where $\mathcal{P}$ denotes the class prototypes initialized via $K$-means clustering, and $\alpha$ is a momentum-based update factor. As shown in Figure~\ref{figure2}, prototype cross-entropy loss $\mathcal{L}_{pce}$ and image-to-text contrastive loss $\mathcal{L}_{i2tce}$ are employed in two domains to enhance representation learning:
\begin{equation}
    \mathcal{L}_{pce}^i=-\log \frac{\exp \left(s\left(I_{i}^\prime, \mathcal{P}_{y_{i}}\right)/\tau\right)}{\sum_{c=1}^{C} \exp \left(s\left(I_{i}^\prime, \mathcal{P}_{c}\right)/\tau\right)},
\end{equation}
and
\begin{equation}
    \mathcal{L}_{i2tce}^i=\sum_{j=1}^{C}-\sigma_j\log \frac{\exp \left(s\left(I_{i}^\prime, T_{j}^ \prime \right)/\tau\right)}{\sum_{c=1}^{C} \exp \left(s\left(I_{i}^\prime, T_{c} ^\prime\right)/\tau\right)},
\end{equation}
where $C$ is the number of the categories. To further mitigate the impact of errors in pseudo-labels, we introduce a smoothing label $\sigma_j=(1-\epsilon)\cdot y_i+\epsilon/C$, where $y_i$ is the one-hot encoding of the pseudo-label and $\epsilon$ is a constant. The overall loss function for TPDP is then computed by summing the individual losses:
\begin{equation}\label{loss}
    \mathcal{L}=\alpha\mathcal{L}_{pce}+\beta\mathcal{L}_{i2tce},
\end{equation}
where $\alpha$ and $\beta$ are hyperparameters balancing the importance of $\mathcal{L}_{pce}$ and $\mathcal{L}_{i2tce}$, respectively.
\section{Experiments}
\label{sec:experiments}
\subsection{Datasets and Setting}

\paragraph{Datasets.}
We evaluate the proposed TPSNet on two widely used datasets for cross-domain retrieval: Office-Home and DomainNet. Office-Home dataset~\cite{officehome} comprises 65 classes and 4 domains (Art, Clipart, Product, Real). DomainNet dataset~\cite{domainnet} consists of 6 domains: Clipart, Infograph, Painting, Quickdraw, Real, and Sketch. Following the criteria established by DD~\cite{ucdir}, we select 7 classes from the DomainNet dataset with over 200 images in each domain for training and testing.

\paragraph{Evaluation Metrics.}
Following DD~\cite{ucdir}, we assess retrieval performance on the Office-Home dataset using precision at top-$k$ retrieved images, specifically at $k = 1$, $5$, and $15$ (i.e., $P@1$, $P@5$, and $P@15$). For the DomainNet dataset, we report precision at $k = 50$, $100$, and $200$ (i.e., $P@50$, $P@100$, and $P@200$).

\begin{table*}[t]
  \caption{Detailed comparison with UCDIR methods on the Office-Home dataset across 12 individual cross-domain retrieval scenarios.}
  \label{table1}

  \centering
  \resizebox{\linewidth}{!}{
    \begin{tabular}{llcccccccccccc}
    \toprule
    \multicolumn{2}{c}{\multirow{2}{*}{Method}} & \multicolumn{3}{c}{$Art \rightarrow Real$} & \multicolumn{3}{c}{$Real \rightarrow Art$} & \multicolumn{3}{c}{$Art \rightarrow Product$} & \multicolumn{3}{c}{$Product \rightarrow Art$} \\ 
    \cmidrule(r){3-5} \cmidrule(r){6-8} \cmidrule(r){9-11} \cmidrule(r){12-14}
    \multicolumn{2}{c}{ }&$P@1$&$P@5$&$P@15$&$P@1$&$P@5$&$P@15$&$P@1$&$P@5$&$P@15$&$P@1$&$P@5$&$P@15$ \\
    \midrule 
    \multirow{4}{*}{ResNet-50}&DD~\cite{ucdir}&45.12&42.33&40.06&47.95&43.68&38.38&35.39&34.67&32.61&42.51&37.94&31.41 \\
    &CoDA~\cite{coda}&44.77&40.99&36.64&44.88&37.54&38.12&34.52&33.96&31.06&40.98&32.24&30.54 \\
    &ProtoOT~\cite{protoot}&49.94&49.24&49.55&54.90&51.81&48.17&44.25&45.95&45.71&53.43&52.07&46.89 \\ 
    &ShieldIR~\cite{shieldir}&51.46 &50.20& 49.27& 57.33& 53.50 &49.26 &44.75& 46.16 &45.33 &56.31& 52.71& 47.39 \\
        \rowcolor{gray!20}&\textbf{TPSNet}&\textbf{69.06}&\textbf{67.22}&\textbf{65.02}&\textbf{74.00}&\textbf{69.72}&\textbf{62.73}&\textbf{59.41}&\textbf{57.21}&\textbf{53.50}&\textbf{63.82}&\textbf{59.98}&\textbf{52.31} \\
    \midrule 
    \multirow{5}{*}{ViT-B}&DD~\cite{ucdir}&60.94&56.44&51.75&61.92&54.14&44.97&58.76&56.55&52.65&60.80&53.89&44.88 \\
    &CoDA~\cite{coda}&59.65&53.68&46.84&58.34&52.82&40.84&57.32&53.86&50.80&57.53&51.54&40.11 \\
    &DG-UCDIR~\cite{dgucdir}&59.29&52.76&46.59&58.83&51.54&39.48&56.51&53.01&52.65&61.46&55.89&47.42 \\
    &SA-MoE~\cite{samoe}&71.12&68.93&66.10&73.86&68.85&60.37&64.69&62.39&58.63&66.57&62.82&55.12 \\ 
        \rowcolor{gray!20}&\textbf{TPSNet}&\textbf{89.53}&\textbf{88.61}&\textbf{87.77}&\textbf{91.74}&\textbf{90.71}&\textbf{86.81}&\textbf{81.17}&\textbf{80.60}&\textbf{79.43}&\textbf{88.80}&\textbf{86.14}&\textbf{81.01} \\
    \midrule 

    \multicolumn{2}{c}{\multirow{2}{*}{Method}} & \multicolumn{3}{c}{$Clipart \rightarrow Real$} & \multicolumn{3}{c}{$Real \rightarrow Clipart$} & \multicolumn{3}{c}{$Product \rightarrow Real$} & \multicolumn{3}{c}{$Real \rightarrow Product$} \\ 
    \cmidrule(r){3-5} \cmidrule(r){6-8} \cmidrule(r){9-11} \cmidrule(r){12-14}
    \multicolumn{2}{c}{ }&$P@1$&$P@5$&$P@15$&$P@1$&$P@5$&$P@15$&$P@1$&$P@5$&$P@15$&$P@1$&$P@5$&$P@15$ \\
    \midrule 
    \multirow{4}{*}{ResNet-50}&DD~\cite{ucdir}&33.31&30.57&28.14&44.66&41.47&37.41&57.42&52.69&47.90&51.71&48.48&44.95 \\
    &CoDA~\cite{coda}&30.12&27.10&24.02&43.65&35.21&29.06&57.37&50.98&42.82&55.23&49.18&44.36 \\
    &ProtoOT~\cite{protoot}&40.92&40.68&39.96&51.11&52.12&50.44&69.34&67.27&64.74&57.72&59.32&59.72 \\ 
    &ShieldIR~\cite{shieldir}&43.36& 42.58& 41.55& 52.05& 52.38 &50.81 &70.37 &68.33 &66.08 &61.46& 62.08 &61.85\\
    \rowcolor{gray!20}&\textbf{TPSNet}&\textbf{51.68}&\textbf{49.91}&\textbf{47.56}&\textbf{63.53}&\textbf{61.36}&\textbf{57.40}&\textbf{78.85}&\textbf{75.87}&\textbf{72.48}&\textbf{74.39}&\textbf{73.00}&\textbf{70.94} \\
    \midrule 
    \multirow{5}{*}{ViT-B}&DD~\cite{ucdir}&42.66&39.51&36.19&59.31&54.58&48.18&70.56&65.19&58.58&65.87&62.57&57.58 \\
    &CoDA~\cite{coda}&41.06&38.46&34.83&57.42&51.47&45.92&70.43&63.62&56.84&63.94&60.13&55.16 \\
    &DG-UCDIR~\cite{dgucdir}&42.87&39.31&35.94&60.86&55.35&48.90&72.12&66.87&60.41&66.25&62.97&58.09 \\
    &SA-MoE~\cite{samoe}&52.99&50.00&47.33&70.41&65.78&60.06&78.91&74.96&70.40&76.52&74.54&71.18 \\ 
        \rowcolor{gray!20}&\textbf{TPSNet}&\textbf{73.36}&\textbf{72.73}&\textbf{71.85}&\textbf{89.03}&\textbf{88.01}&\textbf{84.86}&\textbf{93.26}&\textbf{92.82}&\textbf{91.78}&\textbf{92.10}&\textbf{92.05}&\textbf{91.78} \\
    \midrule 

    \multicolumn{2}{c}{\multirow{2}{*}{Method}} & \multicolumn{3}{c}{$Product \rightarrow Clipart$} & \multicolumn{3}{c}{$Clipart \rightarrow Product$} & \multicolumn{3}{c}{$Art \rightarrow Clipart$} & \multicolumn{3}{c}{$Clipart \rightarrow Art$} \\ 
    \cmidrule(r){3-5} \cmidrule(r){6-8} \cmidrule(r){9-11} \cmidrule(r){12-14}
    \multicolumn{2}{c}{ }&$P@1$&$P@5$&$P@15$&$P@1$&$P@5$&$P@15$&$P@1$&$P@5$&$P@15$&$P@1$&$P@5$&$P@15$ \\
    \midrule 
    \multirow{4}{*}{ResNet-50}&DD~\cite{ucdir}&42.26&37.42&33.74&27.79&27.26&25.97&32.67&30.79&28.70&27.26&23.94&20.53 \\
    &CoDA~\cite{coda}&47.21&35.43&28.33&27.10&24.77&24.00&36.69&32.19&26.37&25.64&21.17&21.37 \\
    &ProtoOT~\cite{protoot}&51.92&50.66&49.58&39.01&37.95&37.16&40.54&38.27&36.01&30.24&29.31&27.34 \\ 
    &ShieldIR~\cite{shieldir}&52.10& 52.43 &50.23 &40.84 &39.87 &39.45 &41.24 &39.05& 37.44& 32.21& 30.13& 28.39\\
    \rowcolor{gray!20}&\textbf{TPSNet}&\textbf{60.71}&\textbf{57.49}&\textbf{53.26}&\textbf{46.05}&\textbf{43.77}&\textbf{42.22}&\textbf{56.98}&\textbf{54.07}&\textbf{50.29}&\textbf{43.14}&\textbf{39.66}&\textbf{35.58} \\
    \midrule 
    \multirow{5}{*}{ViT-B}&DD~\cite{ucdir}&57.87&54.00&47.87&45.22&43.29&40.13&56.00&51.89&45.78&39.06&34.64&29.27\\
    &CoDA~\cite{coda}&54.44&51.97&44.68&44.84&42.62&38.91&55.76&50.41&44.63&38.42&33.02&27.41 \\
    &DG-UCDIR~\cite{dgucdir}&56.58&53.41&47.09&43.89&41.25&37.48&55.93&51.50&44.93&36.51&30.04&25.43 \\
    &SA-MoE~\cite{samoe}&65.28&61.09&56.01&47.47&45.81&43.36&61.80&57.82&53.23&46.90&44.39&39.97 \\ 
        \rowcolor{gray!20}&\textbf{TPSNet}&\textbf{88.31}&\textbf{86.74}&\textbf{84.49}&\textbf{72.67}&\textbf{72.30}&\textbf{71.79}&\textbf{84.55}&\textbf{81.94}&\textbf{79.36}&\textbf{69.87}&\textbf{68.67}&\textbf{64.55} \\
    
    \bottomrule
    \end{tabular}}

\end{table*}
\begin{table*}[!ht]
  \caption{Detailed comparison with UCDIR methods on the DomainNet dataset across 12 individual cross-domain retrieval scenarios.}
  \label{table2}
  \centering
  \resizebox{\linewidth}{!}{
    \begin{tabular}{llcccccccccccc}
    \toprule
    \multicolumn{2}{c}{\multirow{2}{*}{Method}} & \multicolumn{3}{c}{$Clipart \rightarrow Sketch$} & \multicolumn{3}{c}{$Sketch \rightarrow Clipart$} & \multicolumn{3}{c}{$Infograph \rightarrow Real$} & \multicolumn{3}{c}{$Real \rightarrow Infograph$} \\ 
    \cmidrule(r){3-5} \cmidrule(r){6-8} \cmidrule(r){9-11} \cmidrule(r){12-14}
    \multicolumn{2}{c}{ }&$P@50$&$P@100$&$P@200$&$P@50$&$P@100$&$P@200$&$P@50$&$P@100$&$P@200$&$P@50$&$P@100$&$P@200$ \\
    \midrule 
    \multirow{4}{*}{ResNet-50}&DD~\cite{ucdir}&56.31&52.47&47.38&63.07&57.26&48.17&35.52&35.24&34.35& 57.74&46.69&35.47 \\
    &CoDA~\cite{coda}&44.56&35.86&35.14&49.00&38.61&38.49&27.12&27.21&26.43&36.98&30.44&30.02 \\
    &ProtoOT~\cite{protoot}&82.70&82.26&80.32&86.69&85.57&81.59&48.74&49.28&49.60&75.36&68.49&56.20 \\ 
    &ShieldIR~\cite{shieldir}&87.16 &87.12& 86.74 &89.95& 89.54 &88.56 &54.41 &54.93 &55.34 &82.26& 75.35 &64.15\\
    \rowcolor{gray!20}&\textbf{TPSNet}&\textbf{95.43}&\textbf{95.31}&\textbf{94.63}&\textbf{95.50}&\textbf{95.31}&\textbf{94.26}&\textbf{70.04}&\textbf{69.55}&\textbf{68.74}&\textbf{86.07}&\textbf{81.91}&\textbf{71.99}\\
    \midrule 
    \multirow{5}{*}{ViT-B}&DD~\cite{ucdir}&76.55&73.39&68.66&82.36&79.22&71.46&47.12&47.15&46.89 &78.01&68.07&53.17 \\
    &CoDA~\cite{coda}&73.62&70.22 &65.51& 80.21& 76.00& 67.68& 46.11 &45.98 &45.13& 73.69 &61.67 &45.62 \\
    &DG-UCDIR~\cite{dgucdir}&77.47& 73.83& 69.31& 82.86& 80.85& 72.19& 48.92& 47.97& 45.20& 78.72& 68.71& 52.37 \\
    &SA-MoE~\cite{samoe}&83.97& 82.08& 78.40& 88.11& 86.07& 81.21& 57.29 &57.55& 57.67& 87.37& 80.08 &66.24 \\ 
        \rowcolor{gray!20}&\textbf{TPSNet}&\textbf{97.92}&\textbf{97.84}&\textbf{97.73}&\textbf{98.27}&\textbf{98.20}&\textbf{98.06}&\textbf{78.72}&\textbf{78.55}&\textbf{78.31}&\textbf{93.01}&\textbf{93.04}&\textbf{85.71} \\
    \midrule 

    \multicolumn{2}{c}{\multirow{2}{*}{Method}} & \multicolumn{3}{c}{$Infograph \rightarrow Sketch$} & \multicolumn{3}{c}{$Sketch \rightarrow Infograph$} & \multicolumn{3}{c}{$Painting \rightarrow Clipart$} & \multicolumn{3}{c}{$Clipart \rightarrow Painting$} \\ 
    \cmidrule(r){3-5} \cmidrule(r){6-8} \cmidrule(r){9-11} \cmidrule(r){12-14}
    \multicolumn{2}{c}{ }&$P@50$&$P@100$&$P@200$&$P@50$&$P@100$&$P@200$&$P@50$&$P@100$&$P@200$&$P@50$&$P@100$&$P@200$ \\
    \midrule 
    \multirow{4}{*}{ResNet-50}&DD~~\cite{ucdir}&31.29& 29.33& 26.54& 43.66& 36.14& 28.12& 66.42& 56.84 &46.72& 52.58& 50.10 &46.11 \\
    &CoDA~\cite{coda}&24.94& 22.48& 22.42& 27.65& 23.85 &23.48 &57.30& 44.25& 44.15& 46.51& 42.38& 41.22 \\
    &ProtoOT~\cite{protoot}&45.72& 45.51& 43.77& 62.92& 56.41& 47.08& 91.04& 90.09& 86.67& 82.70& 82.66 &81.38 \\ 
    &ShieldIR~\cite{shieldir}&47.39 &47.10 &46.34 &72.29& 67.68 &55.00 &93.96 &\textbf{93.47} &\textbf{90.99} &85.32& 85.34 &84.30\\
    \rowcolor{gray!20}&\textbf{TPSNet}&\textbf{67.59}&\textbf{67.00}&\textbf{65.96}&\textbf{90.20}&\textbf{86.27}&\textbf{74.90}&\textbf{94.07}&93.27&90.61&\textbf{94.56}&\textbf{93.99}&\textbf{91.08} \\
    \midrule 
    \multirow{5}{*}{ViT-B}&DD\cite{ucdir}&49.94& 48.72& 45.11& 71.97& 62.05& 49.32& 93.89& 93.24 &90.09& 86.00& 85.57& 83.72 \\
    &CoDA~\cite{coda}&45.85& 44.52& 40.35& 67.51& 55.69& 43.87& 85.93& 82.34& 80.29& 82.56& 80.18& 79.65 \\
    &DG-UCDIR~\cite{dgucdir}&50.38& 49.24& 45.62& 72.03& 63.15& 51.99& 92.40& 93.17& 89.43& 86.17& 85.83& 82.86 \\
    &SA-MoE~\cite{samoe}&53.76& 53.07& 50.37& 79.56& 72.29& 58.62& 95.00& 94.59& 93.23& 90.60& 90.42& 89.25 \\ 
        \rowcolor{gray!20}&\textbf{TPSNet}&\textbf{75.87}&\textbf{75.61}&\textbf{75.03}&\textbf{95.21}&\textbf{92.04}&\textbf{84.78}&\textbf{98.49}&\textbf{98.45}&\textbf{98.26}&\textbf{97.84}&\textbf{97.73}&\textbf{96.64} \\ 
    \midrule 

    \multicolumn{2}{c}{\multirow{2}{*}{Method}} & \multicolumn{3}{c}{$Painting \rightarrow Quickdraw$} & \multicolumn{3}{c}{$Quickdraw \rightarrow Painting$} & \multicolumn{3}{c}{$Quickdraw \rightarrow Real$} & \multicolumn{3}{c}{$Real \rightarrow Quickdraw$} \\ 
    \cmidrule(r){3-5} \cmidrule(r){6-8} \cmidrule(r){9-11} \cmidrule(r){12-14}
    \multicolumn{2}{c}{ }&$P@50$&$P@100$&$P@200$&$P@50$&$P@100$&$P@200$&$P@50$&$P@100$&$P@200$&$P@50$&$P@100$&$P@200$ \\
    \midrule 
    \multirow{4}{*}{ResNet-50}&DD~\cite{ucdir}&39.72& 38.59 &37.63& 33.45& 33.81& 34.29& 30.24& 28.40& 26.14& 25.00& 22.26& 19.73 \\
    &CoDA~\cite{coda}&28.73& 26.69& 24.95& 21.05& 20.52& 20.81& 23.98& 22.59& 21.50& 39.87& 36.28& 32.91 \\
    &ProtoOT~\cite{protoot}&63.06& 59.92& 56.73& 56.51&55.66& 52.04& 61.55& 61.60& 61.63& 67.24& 67.02 &65.41 \\ 
    &ShieldIR~\cite{shieldir}&65.33 &61.25& 57.98& 58.07 &57.42& \textbf{55.11}& 67.19 &67.36 &67.53& \textbf{75.70}& 73.60 &70.63\\
        \rowcolor{gray!20}&\textbf{TPSNet}&\textbf{67.21}&\textbf{65.38}&\textbf{61.88}&\textbf{59.27}&\textbf{57.93}&54.30&\textbf{69.34}&\textbf{68.88}&\textbf{68.17}&75.06&\textbf{74.70}&\textbf{72.96} \\
    \midrule 
    \multirow{5}{*}{ViT-B}&DD~\cite{ucdir}&63.19& 61.19& 58.09& 53.55& 53.46& 53.09 &48.71& 48.74 &48.68& 51.65& 50.48& 48.97\\
    &CoDA~\cite{coda}&50.24& 48.58& 44.68& 43.49& 42.65& 41.94& 37.67& 29.63& 26.21& 39.85& 31.84& 29.49 \\
    &DG-UCDIR~\cite{dgucdir}&65.06& 62.90& 59.89& 54.18& 53.95& 53.21& 45.20& 44.30& 42.63& 50.62& 49.24& 46.80 \\
    &SA-MoE~\cite{samoe}&72.54& 71.16&69.22&61.32&61.21&61.03& 55.78& 55.59& 55.49& 59.02& 57.55& 55.81 \\ 
        \rowcolor{gray!20}&\textbf{TPSNet}&\textbf{81.67}&\textbf{77.04}&\textbf{71.34}&\textbf{65.12}&\textbf{64.51}&\textbf{63.31}&\textbf{62.50}&\textbf{62.21}&\textbf{61.56}&\textbf{79.34}&\textbf{76.89}&\textbf{72.14} \\
    \bottomrule
    \end{tabular}}
\end{table*}

\paragraph{Implementation Details.}
We use the pre-trained CLIP~\cite{clip} architecture as the backbone framework. Following the consensus in this community, we adopt ResNet-50~\cite{resnet} and ViT-B~\cite{vit} as image encoder to investigate the effectiveness of TPSNet. During the DPG stage, the CLIP backbone is frozen, and only the learnable text tokens $\left[X\right]^1\left[X\right]^2...\left[X\right]^M$ are updated. In the TPDP stage, we unfreeze the last block of the image encoder, allowing fine-tuning. Each input image is augmented using random horizontal flipping, padding, cropping, and random erasing. The model is optimized using the Adam optimizer with a learning rate of 0.0001, following a cosine decay schedule. All the experiments are conducted on a single NVIDIA GeForce RTX 3090.
\subsection{Comparison with State-of-the-Art Methods}
\begin{figure}[h!]
  \centering
  \includegraphics[width=1\linewidth]{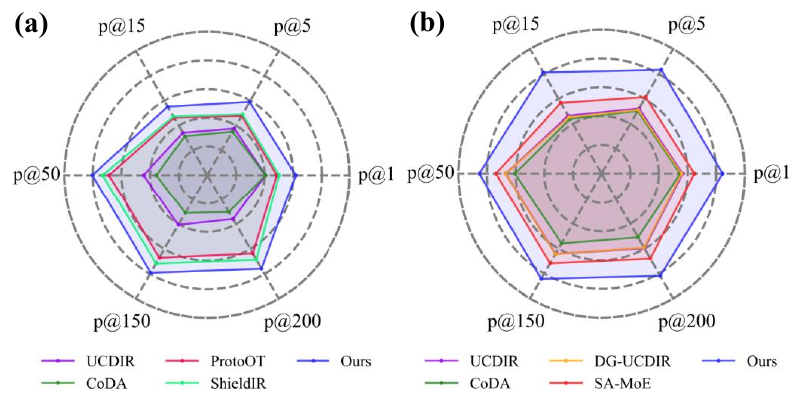}
  \vspace{-1em}
  \caption{Average Accuracy (\%) of UCDIR Methods using (a) ResNet-50 and (b) ViT-B as image encoders.}
  \label{figure3}
\end{figure}
In our experimental validation, we evaluate the effectiveness of TPSNet by comparing it with the following methods: DD~\cite{ucdir}, CoDA~\cite{coda}, DG-UCDIR~\cite{dgucdir}, ProtoOT~\cite{protoot}, ShieldIR~\cite{shieldir} and SA-MoE~\cite{samoe}. Table~\ref{table1} and Table~\ref{table2} show the detailed retrieval performance for Office-Home and DomainNet, respectively. TPSNet's impressive performance is evident from the summarized results in Figure~\ref{figure3}, with detailed average accuracies provided in Table~\ref{table3}.

\begin{table}[!ht]
  \caption{Average accuracy (\%) for evaluating the impact of image encoder initialization on the Office-Home and DomainNet datasets using ResNet-50 and ViT-B backbones.}
  \label{table3}
  \resizebox{\linewidth}{!}{

    \begin{tabular}{lcccccc}
    \toprule
    \multirow{3}{*}{Method} & \multicolumn{6}{c}{ResNet-50}  \\ 
    \cmidrule(r){2-7} 
    \multicolumn{1}{c}{} & \multicolumn{3}{c}{Office-Home} & \multicolumn{3}{c}{Domainnet}  \\ 
    \cmidrule(r){2-4} \cmidrule(r){5-7} 
    &$P@1$&$P@5$&$P@15$&$P@50$&$P@100$&$P@200$\\   \cmidrule(r){1-2}\cmidrule(r){2-4} \cmidrule(r){5-7} 
    SOTA~\cite{shieldir}&50.29&49.12&47.25
&73.25&71.68&68.56 \\ 
    TPSNet (MoCov2)&56.71&54.64&52.14&74.04&72.80&70.00 \\ 
        \rowcolor{gray!20}\textbf{TPSNet}&\textbf{61.80}&\textbf{59.11}&\textbf{55.27}&\textbf{80.36}&\textbf{79.13}&\textbf{75.79}\\ 
    \cmidrule(r){1-7}
\multirow{3}{*}{Method} & \multicolumn{6}{c}{ViT-B}  \\ 
    \cmidrule(r){2-7} 
    \multicolumn{1}{c}{} & \multicolumn{3}{c}{Office-Home} & \multicolumn{3}{c}{Domainnet}  \\ 
    \cmidrule(r){2-4} \cmidrule(r){5-7} 
    &$P@1$&$P@5$&$P@15$&$P@50$&$P@100$&$P@200$\\   \cmidrule(r){1-2}\cmidrule(r){2-4} \cmidrule(r){5-7} 
    SOTA~\cite{samoe}&64.71&61.45&56.81&73.69&71.81&68.05 \\ 
    TPSNet (DINO)&67.62&65.31&62.29&80.97&79.30&75.48 \\ 
        \rowcolor{gray!20}\textbf{TPSNet}&\textbf{84.53}&\textbf{83.44}&\textbf{81.29}&\textbf{85.33}&\textbf{84.34}&\textbf{81.91}\\ 
            \bottomrule
    \vspace{-3em}
\end{tabular}}
\end{table}

Through comparison, we observe the following: 1) TPSNet outperforms other methods on nearly all metrics across the two datasets, highlighting its effectiveness. 2) With either ResNet-50 or ViT-B as the image encoder, TPSNet outperforms other compared methods on nearly all metrics, demonstrating its robustness. 3) TPSNet exhibits more substantial performance gains on the Office-Home dataset, which contains a larger number of categories (65 in total). Specifically, when using ResNet-50 as the image encoder, TPSNet surpasses the SOTA by average margins of 11.51\%, 9.99\%, and 8.02\% in $P@1$, $P@5$, and $P@15$, respectively. The improvements are even more pronounced with ViT-B as the image encoder, yielding gains of 19.82\%, 22.00\%, and 24.48\% in the same metrics. These results highlight the effectiveness of TPSNet in handling more complex and challenging datasets. 4) We observe that TPSNet achieves substantial improvements on noisy domains with large visual variation. For instance, on the Infograph-real and Infograph-sketch scenarios of the DomainNet dataset, TPSNet outperforms SOTA by more than 20\% in $P@50$ when ViT-B is used as the image encoder. These results demonstrate that TPSNet facilitates more effective semantic learning, thereby improving the model’s robustness to the domain distribution gaps. 

In addition, to comprehensively evaluate the effectiveness and robustness of TPSNet and ensure that the observed improvements are not solely attributed to CLIP, we replace the pre-trained CLIP image encoder with MoCov2~\cite{moco} and DINO~\cite{DINO}, which are widely adopted in prior works~\cite{ucdir,protoot,samoe}. All backbones are trained under identical settings for fair comparison. As shown in Table~\ref{table3}, TPSNet consistently outperforms state-of-the-art methods across both datasets, even when MoCov2 and DINO are used for initialization. These results strongly indicate that the performance gains are not merely dependent on CLIP, but instead reflect the intrinsic effectiveness and generalizability of TPSNet. Additional detailed comparisons are provided in Supplementary Material B.4.
\subsection{Ablation Study}
\paragraph{Quantitative Evaluation.}
To evaluate the contribution of each component in TPSNet, we perform an ablation study on two datasets using ViT-B as the image encoder, with results summarized in Table~\ref{table4}. The results indicate that the integration of domain prompts provides substantial performance gains by introducing additional semantic supervision, yielding an improvement of 11.91\% in $P@1$ on the Office-Home dataset and 6.69\% in $P@50$ on DomainNet. Furthermore, the inclusion of TPFE and PPFE modules leads to additional performance enhancements by strengthening semantic information while reducing domain discrepancies. When all components are combined, TPSNet achieves further performance gains, proving the efficacy of our method.

\begin{table}[h]
  \centering
  \caption{Ablation study of TPSNet components with ViT-B as the image encoder. Results averaged over 12 scenarios.}
  \label{table4}
  \setlength{\tabcolsep}{3pt}
  \resizebox{\linewidth}{!}{
  \begin{tabular}{ccccccccc}
    \toprule
    \multirow{2}{*}{DPG} & \multirow{2}{*}{TPFE} & \multirow{2}{*}{PPFE} 
    & \multicolumn{3}{c}{Office-Home} & \multicolumn{3}{c}{DomainNet} \\ 
    \cmidrule(r){4-6} \cmidrule(r){7-9} 
    &&& $P@1$ & $P@5$ & $P@15$ & $P@50$ & $P@100$ & $P@200$ \\ 
    \midrule 
    \XSolidBrush & \XSolidBrush & \XSolidBrush & 65.18 & 60.77 & 54.87 & 70.94 & 68.09 & 62.58 \\ 
    \Checkmark   & \XSolidBrush & \XSolidBrush & 77.09 & 75.10 & 71.43 & 77.63 & 74.72 & 69.18 \\ 
    \Checkmark   & \XSolidBrush & \Checkmark   & 81.34 & 80.22 & 78.07 & 81.88 & 79.73 & 75.75 \\ 
    \Checkmark   & \Checkmark   & \XSolidBrush & 81.63 & 80.13 & 77.72 & 82.05 & 79.94 & 75.97 \\ 
    \rowcolor{gray!20}
    \Checkmark   & \Checkmark   & \Checkmark   & \textbf{84.53} & \textbf{83.44} & \textbf{81.29} & \textbf{85.33} & \textbf{84.34} & \textbf{81.91} \\ 
    \bottomrule 
  \end{tabular}
  }
\end{table}
\begin{figure}[h]
    \centering
    \includegraphics[width=1\linewidth]{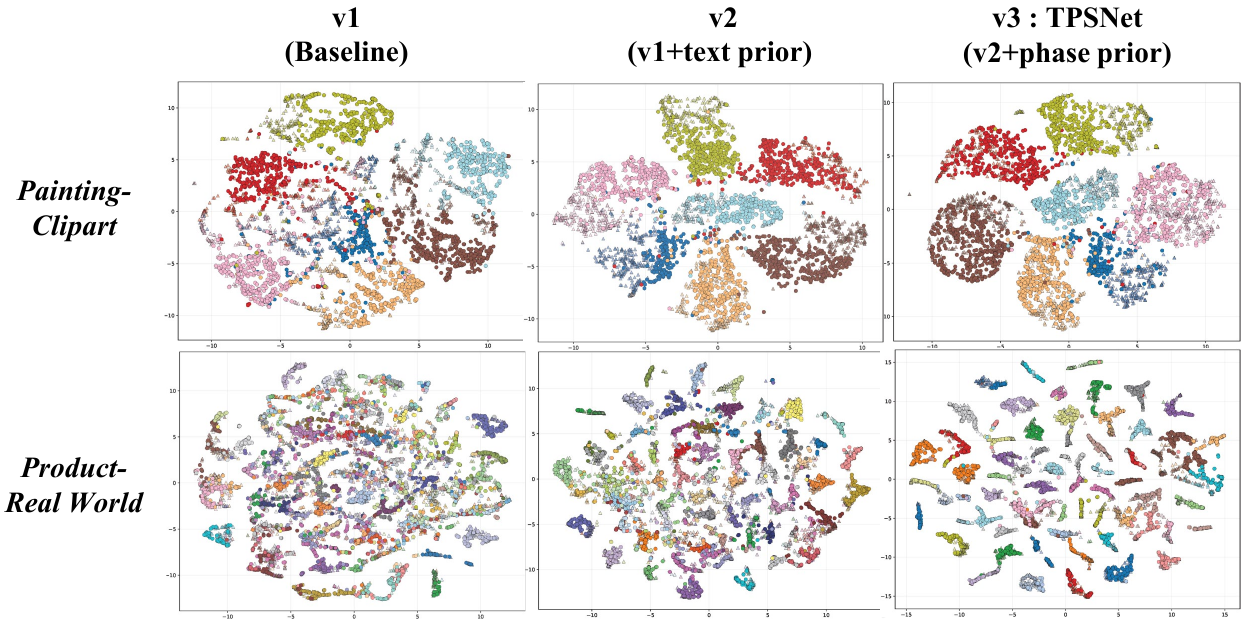}
    \caption{t-SNE visualizations of last-layer features for the baseline model (v1), baseline with text prior (v2), and TPSNet (v3) across two scenarios from two datasets.}
    \label{figure4}
\end{figure}

\begin{figure}[h]
    \centering
    \vspace{-1em}
    \includegraphics[width=1\linewidth]{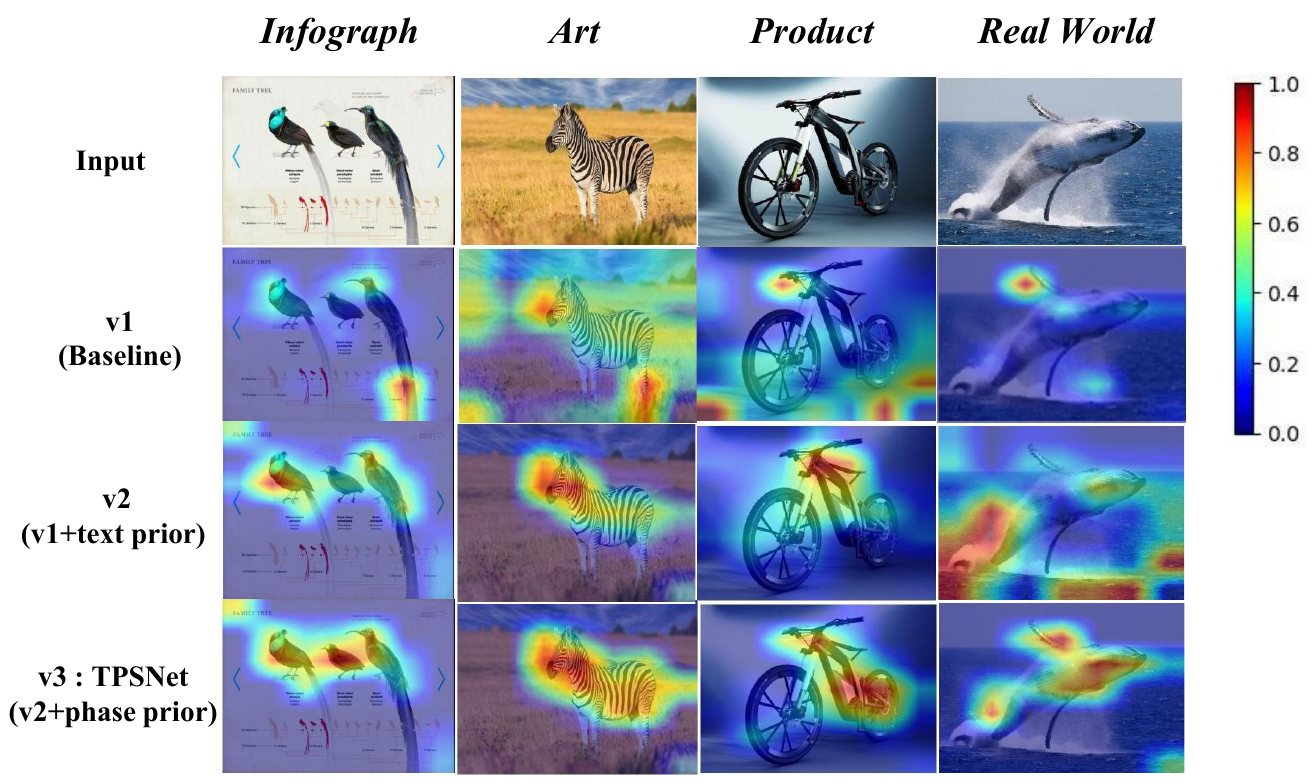}
    \caption{Grad-CAM visualizations of last-layer features for the baseline model (v1), baseline with text prior (v2), and TPSNet (v3) on randomly selected samples.}
    \label{figure5}
\end{figure}
\paragraph{Qualitative Evaluation.}


To evaluate the explainability of the proposed text-phase dual priors in semantic feature extraction and domain distribution alignment, we perform two sets of qualitative assessments. Specifically, we compare three models: the baseline model (v1), the baseline model enhanced with the text prior (v2), and TPSNet (v3), which integrates all components.

First, we visualize the t-SNE~\cite{tsne} representations of features extracted from the final layer of the image encoder, as shown in Figure~\ref{figure4}. The results show that incorporating the text prior facilitates more compact clustering of semantically similar samples, while the phase prior further aligns similar samples across domains into a shared feature space. This highlights the effectiveness of TPSNet in learning domain-invariant semantic representations. Subsequently, we present Grad-CAM~\cite{GradCam} visualizations of the last-layer features from the backbone for models v1, v2, and v3 in Figure~\ref{figure5}. The results indicate that the synergy of the text-phase dual priors enables TPSNet to more accurately focus on target objects, such as the bird in a complex background and the whale leaping across the sea surface. More visualizations can be found in Supplementary Material B.9.


\begin{figure}[t]
    \centering
    \includegraphics[width=1\linewidth]{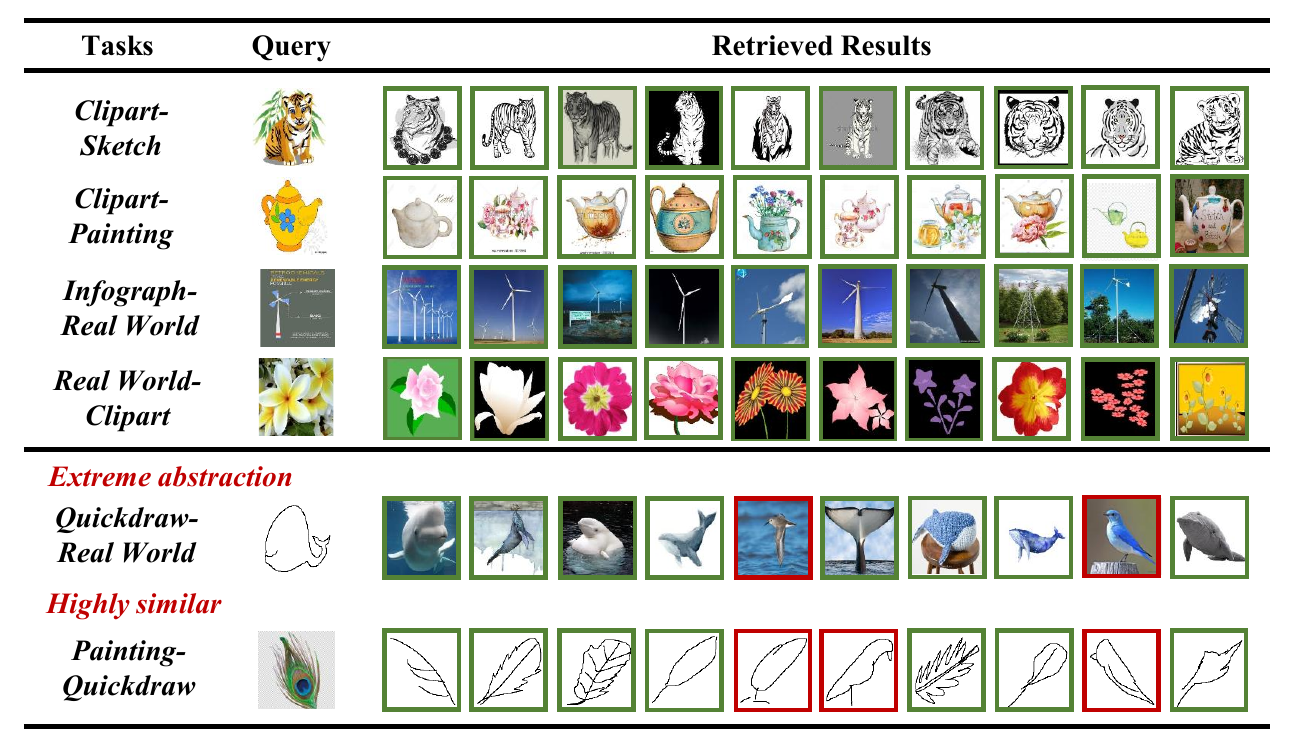}
    \caption{Top-10 retrieval results in various cross-domain retrieval scenarios from two datasets. The green and red boxes denote correct and incorrect retrievals, respectively.}
    \vspace{-1em}
    \label{figure6}
\end{figure}

\paragraph{Visualization of the Retrieval Results.}
We randomly select various cross-domain retrieval scenarios from two datasets for evaluation. As shown in Figure~\ref{figure6}, TPSNet achieves 100\% top-10 retrieval accuracy in most scenarios. We further analyze the failure cases and find that retrieval errors mainly arise under two conditions:
1) queries come from highly abstract domains such as Quickdraw, where fine-grained semantic information is severely limited, and 2) query and retrieved samples exhibit strong visual similarity, for example when bird and feather instances become nearly indistinguishable in the Quickdraw domain.

\section{Conclusion}
\label{sec:conclusion}
In this paper, we address the challenge of unsupervised cross-domain image retrieval (UCDIR) by proposing a novel method called TPSNet. TPSNet leverages domain prompts derived from image-text multi-modal alignment as a text prior to refine inaccurate pseudo-labels and enhance semantic representations. In addition, it integrates phase features as a phase prior to reduce domain discrepancies while preserving semantic information. Through the synergy of text-phase dual priors, TPSNet effectively extracts domain-invariant semantic representations for UCDIR. Extensive experiments on two datasets demonstrate its significant superiority over state-of-the-art methods. 
{
    \small
    \bibliographystyle{ieeenat_fullname}
    \bibliography{main}

@String(AAAI = {AAAI})

@inproceedings{mobile,
  title={Mobile product image search by automatic query object extraction},
  author={Shen, Xiaohui and Lin, Zhe and Brandt, Jonathan and Wu, Ying},
  booktitle={Proceedings of the European Conference on Computer Vision},
  pages={114--127},
  year={2012},
  publisher={Springer}
}

@article{face,
  title={Face recognition using content based image retrieval for intelligent security},
  author={Karnila, Sri and Irianto, Suhendro and Kurniawan, Rio},
  journal={International Journal of Advanced Engineering Research and Science},
  volume={6},
  number={1},
  pages={91--98},
  year={2019}
}

@inproceedings{medical,
  title={X-mir: Explainable medical image retrieval},
  author={Hu, Brian and Vasu, Bhavan and Hoogs, Anthony},
  booktitle={Proceedings of the IEEE Winter Conference on Applications of Computer Vision},
  pages={440--450},
  year={2022}
}

@inproceedings{cdir1,
  title={Cross-domain image retrieval with a dual attribute-aware ranking network},
  author={Huang, Junshi and Feris, Rogerio S and Chen, Qiang and Yan, Shuicheng},
  booktitle={Proceedings of the IEEE International Conference on Computer Vision},
  pages={1062--1070},
  year={2015}
}

@inproceedings{cdir2,
  title={Cross-domain image retrieval with attention modeling},
  author={Ji, Xin and Wang, Wei and Zhang, Meihui and Yang, Yang},
  booktitle={Proceedings of the 25th ACM International Conference on Multimedia},
  pages={1654--1662},
  year={2017}
}

@inproceedings{cdir4,
  title={Boosting Fine-grained Fashion Retrieval with Relational Knowledge Distillation},
  author={Xiao, Ling and Yamasaki, Toshihiko},
  booktitle={Proceedings of the IEEE Conference on Computer Vision and Pattern Recognition},
  pages={8229--8234},
  year={2024}
}

@inproceedings{cdir5,
  title={Pros: Prompting-to-simulate generalized knowledge for universal cross-domain retrieval},
  author={Fang, Kaipeng and Song, Jingkuan and Gao, Lianli and Zeng, Pengpeng and Cheng, Zhi-Qi and Li, Xiyao and Shen, Heng Tao},
  booktitle={Proceedings of the IEEE Conference on Computer Vision and Pattern Recognition},
  pages={17292--17301},
  year={2024}
}

@inproceedings{ucdir,
  title={Feature representation learning for unsupervised cross-domain image retrieval},
  author={Hu, Conghui and Lee, Gim Hee},
  booktitle={Proceedings of the European Conference on Computer Vision},
  pages={529--544},
  year={2022},
  organization={Springer}
}

@inproceedings{coda,
  title={Correspondence-free domain alignment for unsupervised cross-domain image retrieval},
  author={Wang, Xu and Peng, Dezhong and Yan, Ming and Hu, Peng},
  booktitle={Proceedings of the AAAI Conference on Artificial Intelligence},
  pages={10200--10208},
  year={2023}
}

@inproceedings{dgucdir,
  title={Unsupervised feature representation learning for domain-generalized cross-domain image retrieval},
  author={Hu, Conghui and Zhang, Can and Lee, Gim Hee},
  booktitle={Proceedings of the IEEE International Conference on Computer Vision},
  pages={11016--11025},
  year={2023}
}

@inproceedings{protoot,
  title={Unsupervised cross-domain image retrieval via prototypical optimal transport},
  author={Li, Bin and Shi, Ye and Yu, Qian and Wang, Jingya},
  booktitle={Proceedings of the AAAI Conference on Artificial Intelligence},
  pages={3009--3017},
  year={2024}
}

@inproceedings{samoe,
  title={Unsupervised Cross-Domain Image Retrieval with Semantic-Attended Mixture-of-Experts},
  author={Wang, Kai and Liu, Jiayang and Xu, Xing and Song, Jingkuan and Liu, Xin and Shen, Heng Tao},
  booktitle={Proceedings of the 47th International ACM SIGIR Conference on Research and Development in Information Retrieval},
  pages={197--207},
  year={2024}
}

@article{DA2,
  title={Domain adaptation via prompt learning},
  author={Ge, Chunjiang and Huang, Rui and Xie, Mixue and Lai, Zihang and Song, Shiji and Li, Shuang and Huang, Gao},
  journal={IEEE Transactions on Neural Networks and Learning Systems},
  volume={36},
  number={1},
  pages={1160-1170},
  year={2023},
  publisher={IEEE}
}

@inproceedings{clip,
  title={Learning transferable visual models from natural language supervision},
  author={Radford, Alec and Kim, Jong Wook and Hallacy, Chris and Ramesh, Aditya and Goh, Gabriel and Agarwal, Sandhini and Sastry, Girish and Askell, Amanda and Mishkin, Pamela and Clark, Jack and others},
  booktitle={Proceedings of the International Conference on Machine Learning},
  pages={8748--8763},
  year={2021},
  organization={PMLR}
}

@inproceedings{Styleclip,
  title={Styleclip: Text-driven manipulation of stylegan imagery},
  author={Patashnik, Or and Wu, Zongze and Shechtman, Eli and Cohen-Or, Daniel and Lischinski, Dani},
  booktitle={Proceedings of the IEEE International Conference on Computer Vision},
  pages={2085--2094},
  year={2021}
}

@inproceedings{Denseclip,
  title={Denseclip: Language-guided dense prediction with context-aware prompting},
  author={Rao, Yongming and Zhao, Wenliang and Chen, Guangyi and Tang, Yansong and Zhu, Zheng and Huang, Guan and Zhou, Jie and Lu, Jiwen},
  booktitle={Proceedings of the IEEE Conference on Computer Vision and Pattern Recognition},
  pages={18082--18091},
  year={2022}
}

@inproceedings{prompt,
  title={Prompt-based distribution alignment for unsupervised domain adaptation},
  author={Bai, Shuanghao and Zhang, Min and Zhou, Wanqi and Huang, Siteng and Luan, Zhirong and Wang, Donglin and Chen, Badong},
  booktitle={Proceedings of the AAAI conference on Artificial Intelligence},
  pages={729--737},
  year={2024}
}

@article{tsne,
  title={Visualizing data using t-SNE},
  author={Van der Maaten, Laurens and Hinton, Geoffrey},
  journal={Journal of Machine Learning Research},
  volume={9},
  number={86},
  pages={2579--2605},
  year={2008}
}

@inproceedings{cdir6,
  title={Dida: Disambiguated domain alignment for cross-domain retrieval with partial labels},
  author={Liu, Haoran and Ma, Ying and Yan, Ming and Chen, Yingke and Peng, Dezhong and Wang, Xu},
  booktitle={Proceedings of the AAAI conference on Artificial Intelligence},
  pages={3612--3620},
  year={2024}
}

@inproceedings{semantic,
    title={Semantic Feature Learning for Universal Unsupervised Cross-Domain Retrieval},
    author={Lixu Wang and Xinyu Du and Qi Zhu},
    booktitle={Proceedings of the Annual Conference on Neural Information Processing Systems},
    year={2024}
}

@inproceedings{vilt,
  title={Vilt: Vision-and-language transformer without convolution or region supervision},
  author={Kim, Wonjae and Son, Bokyung and Kim, Ildoo},
  booktitle={Proceedings of the International Conference on Machine Learning},
  pages={5583--5594},
  year={2021}
}

@inproceedings{VLR1,
  title={Scaling up visual and vision-language representation learning with noisy text supervision},
  author={Jia, Chao and Yang, Yinfei and Xia, Ye and Chen, Yi-Ting and Parekh, Zarana and Pham, Hieu and Le, Quoc and Sung, Yun-Hsuan and Li, Zhen and Duerig, Tom},
  booktitle={Proceedings of the International Conference on Machine Learning},
  pages={4904--4916},
  year={2021}
}

@inproceedings{resnet,
  title={Deep residual learning for image recognition},
  author={He, Kaiming and Zhang, Xiangyu and Ren, Shaoqing and Sun, Jian},
  booktitle={Proceedings of the IEEE Conference on Computer Vision and Pattern Recognition},
  pages={770--778},
  year={2016}
}

@inproceedings{vit,
  title={An Image is Worth 16x16 Words: Transformers for Image Recognition at Scale},
  author={Dosovitskiy, Alexey and Beyer, Lucas and Kolesnikov, Alexander and Weissenborn, Dirk and Zhai, Xiaohua and Unterthiner, Thomas and Dehghani, Mostafa and Minderer, Matthias and Heigold, G and Gelly, S and others},
  booktitle={Proceedings of the International Conference on Learning Representations},
  year={2020}
}

@inproceedings{officehome,
  title={Deep hashing network for unsupervised domain adaptation},
  author={Venkateswara, Hemanth and Eusebio, Jose and Chakraborty, Shayok and Panchanathan, Sethuraman},
  booktitle={Proceedings of the IEEE Conference on Computer Vision and Pattern Recognition},
  pages={5018-5027},
  year={2017}
}

@inproceedings{domainnet,
  title={Moment matching for multi-source domain adaptation},
  author={Peng, Xingchao and Bai, Qinxun and Xia, Xide and Huang, Zijun and Saenko, Kate and Wang, Bo},
  booktitle={Proceedings of the IEEE International Conference on Computer Vision},
  pages={1406-1415},
  year={2019}
}

@inproceedings{GradCam,
  title={Grad-CAM: Visual Explanations from Deep Networks via Gradient-Based Localization},
  author={Selvaraju, Ramprasaath R.  and  Cogswell, Michael  and  Das, Abhishek  and  Vedantam, Ramakrishna  and  Parikh, Devi  and  Batra, Dhruv },
  booktitle={Proceedings of the IEEE International Conference on Computer Vision},
  pages={618-626},
  year={2017}
}

@article{moco,
  title={Improved baselines with momentum contrastive learning},
  author={Chen, Xinlei and Fan, Haoqi and Girshick, Ross and He, Kaiming},
  journal={arXiv preprint arXiv:2003.04297},
  year={2020}
}

@inproceedings{DINO,
  title={Emerging properties in self-supervised vision transformers},
  author={Caron, Mathilde and Touvron, Hugo and Misra, Ishan and J{\'e}gou, Herv{\'e} and Mairal, Julien and Bojanowski, Piotr and Joulin, Armand},
  booktitle={Proceedings of the IEEE International Conference on Computer Vision},
  pages={9650--9660},
  year={2021}
}

@inproceedings{MMD,
  title={Do we really need to access the source data? source hypothesis transfer for unsupervised domain adaptation},
  author={Liang, Jian and Hu, Dapeng and Feng, Jiashi},
  booktitle={Proceedings of the International Conference on Machine Learning},
  pages={6028--6039},
  year={2020}
}

@inproceedings{MDD,
  title={Bridging Theory and Algorithm for Domain Adaptation},
  author={Zhang, Yuchen and Liu, Tianle and Long, Mingsheng and Jordan, Michael},
  booktitle={Proceedings of the 36th International Conference on Machine Learning},
  pages={7404--7413},
  year={2019}
}

@article{DANN,
  title={Domain-adversarial training of neural networks},
  author={Ganin, Yaroslav and Ustinova, Evgeniya and Ajakan, Hana and Germain, Pascal and Larochelle, Hugo and Laviolette, Fran{\c{c}}ois and March, Mario and Lempitsky, Victor},
  journal={Journal of Machine Learning Research},
  volume={17},
  number={59},
  pages={1--35},
  year={2016}
}

@inproceedings{CDAN,
  title = {Conditional Adversarial Domain Adaptation},
  author = {Long, Mingsheng and CAO, ZHANGJIE and Wang, Jianmin and Jordan, Michael I},
  booktitle = {Proceedings of the Annual Conference on Neural Information Processing Systems},
  pages = {},
  year = {2018}
}

@inproceedings{OT1,
  title={Self-labelling via simultaneous clustering and representation learning},
  author={Asano, YM and Rupprecht, C and Vedaldi, A},
  booktitle={Proceedings of the International Conference on Learning Representations},
  year ={2019}
}

@inproceedings{OT2,
  title={Unsupervised learning of visual features by contrasting cluster assignments},
  author={Caron, Mathilde and Misra, Ishan and Mairal, Julien and Goyal, Priya and Bojanowski, Piotr and Joulin, Armand},
  booktitle = {Proceedings of the Annual Conference on Neural Information Processing Systems},
  pages={9912--9924},
  year={2020}
}

@inproceedings{FDA,
  title={Fda: Fourier domain adaptation for semantic segmentation},
  author={Yang, Yanchao and Soatto, Stefano},
  booktitle={Proceedings of the IEEE Conference on Computer Vision and Pattern Recognition},
  pages={4085--4095},
  year={2020}
}

@article{FUDA,
  title={Boosting unsupervised domain adaptation: A Fourier approach},
  author={Wang, Mengzhu and Wang, Shanshan and Wang, Ye and Wang, Wei and Liang, Tianyi and Chen, Junyang and Luo, Zhigang},
  journal={Knowledge-Based Systems},
  volume={264},
  pages={110325},
  year={2023},
  publisher={Elsevier}
}

@inproceedings{blip,
  title={Blip: Bootstrapping language-image pre-training for unified vision-language understanding and generation},
  author={Li, Junnan and Li, Dongxu and Xiong, Caiming and Hoi, Steven},
  booktitle={Proceedings of the International Conference on Machine Learning},
  pages={12888--12900},
  year={2022}
}

@inproceedings{siglip,
  title={Sigmoid loss for language image pre-training},
  author={Zhai, Xiaohua and Mustafa, Basil and Kolesnikov, Alexander and Beyer, Lucas},
  booktitle={Proceedings of the IEEE International Conference on Computer Vision},
  pages={11975--11986},
  year={2023}
}

@inproceedings{lamra,
  title={LamRA: Large Multimodal Model as Your Advanced Retrieval Assistant}, 
  author={Yikun Liu and Pingan Chen and Jiayin Cai and Xiaolong Jiang and Yao Hu and Jiangchao Yao and Yanfeng Wang and Weidi Xie},
  booktitle={Proceedings of the IEEE Conference on Computer Vision and Pattern Recognition}, 
  year={2025}
}

@inproceedings{mmembed,
  title={Mm-embed: Universal multimodal retrieval with multimodal llms},
  author={Lin, Sheng-Chieh and Lee, Chankyu and Shoeybi, Mohammad and Lin, Jimmy and Catanzaro, Bryan and Ping, Wei},
  booktitle={Proceedings of the International Conference on Learning Representations},
  year={2025}
}

@article{siglip2,
  title={Siglip 2: Multilingual vision-language encoders with improved semantic understanding, localization, and dense features},
  author={Tschannen, Michael and Gritsenko, Alexey and Wang, Xiao and Naeem, Muhammad Ferjad and Alabdulmohsin, Ibrahim and Parthasarathy, Nikhil and Evans, Talfan and Beyer, Lucas and Xia, Ye and Mustafa, Basil and others},
  journal={arXiv preprint arXiv:2502.14786},
  year={2025}
}

@article{phase,
  title={The importance of phase in signals},
  author={Oppenheim, Alan V and Lim, Jae S},
  journal={Proceedings of the IEEE},
  volume={69},
  number={5},
  pages={529--541},
  year={1981}
}

@article{qwen2.5,
  title={Qwen2. 5-vl technical report},
  author={Bai, Shuai and Chen, Keqin and Liu, Xuejing and Wang, Jialin and Ge, Wenbin and Song, Sibo and Dang, Kai and Wang, Peng and Wang, Shijie and Tang, Jun and others},
  journal={arXiv preprint arXiv:2502.13923},
  year={2025}
}

@article{sail,
  title={SAIL-VL2 Technical Report},
  author={Yin, Weijie and Ye, Yongjie and Shu, Fangxun and Liao, Yue and Kang, Zijian and Dong, Hongyuan and Yu, Haiyang and Yang, Dingkang and Wang, Jiacong and Wang, Han and others},
  journal={arXiv preprint arXiv:2509.14033},
  year={2025}
}

@article{internvl,
  title={Internvl3. 5: Advancing open-source multimodal models in versatility, reasoning, and efficiency},
  author={Wang, Weiyun and Gao, Zhangwei and Gu, Lixin and Pu, Hengjun and Cui, Long and Wei, Xingguang and Liu, Zhaoyang and Jing, Linglin and Ye, Shenglong and Shao, Jie and others},
  journal={arXiv preprint arXiv:2508.18265},
  year={2025}
}

@inproceedings{shieldir,
  title={ShieldIR: Privacy-Preserving Unsupervised Cross-Domain Image Retrieval via Dual Protection Transformation},
  author={Tang, Zixin and Fan, Haihui and Zhang, Jinchao and Ma, Hui and Gu, Xiaoyan and Li, Bo and Wang, Weiping},
  booktitle={Proceedings of the 33rd ACM International Conference on Multimedia},
  pages={6383--6392},
  year={2025}
}
}

\appendix
\clearpage
\setcounter{page}{1}
\maketitlesupplementary

In the technical appendices and supplementary material, we provide:
\begin{itemize}
\item Additional implementation details of the proposed methods.
\item Further experimental results and ablation studies.
\item A broader discussion of the limitations of this work.
\end{itemize}
\setcounter{section}{0}
\renewcommand{\thesection}{\Alph{section}} 
\section{Implementation Details}\label{sec:appendix A}        

\subsection{Re-Paired Image and Text Embeddings.}
In DPG, the loss in Eq.(\ref{eq:1}) and Eq.(\ref{eq:2}) are computed between re-paired image and text embeddings. The re-pairing process is performed dynamically during each training iteration based on the cosine similarity, defined as: 
\begin{equation}
s(I_i,T_c) = \frac{I_i^{\top} T_c}{\|I_i\| \|T_c\|}.
\end{equation}
Here, $I_i$ denotes the image embedding, and $T_c$ represents the $c$-th domain prompt embedding, where $c \in {1, \dots, C}$ and $C$ is the number of domain prompts.
Subsequently, the image embedding is paired with the text embedding corresponding to the domain prompt with the highest similarity:
\begin{equation}
y_i = \arg\max_{c} s(I_i,T_c), \quad T_i = T_{y_i}.
\end{equation}
Thus, the re-paired image-text pair $(I_i, T_i)$ is dynamically updated at each iteration to maximize semantic consistency. This strategy enhances the quality of pseudo-labels and facilitates improved semantic representation learning despite domain discrepancies.

\subsection{Phase Feature Encoder.}

In PPFE, we utilize the Phase Feature Encoder to extract phase features. Specifically, after applying the Fast Fourier Transform (FFT), we retain the phase spectrum while replacing the amplitude spectrum with a constant value $R$, which helps preserve domain-invariant information. To investigate the influence of the constant value on the reconstructed image, we analyze the Inverse Fast Fourier Transform (IFFT) of a complex signal constructed as $R \cdot e^{j\phi(u,v)}$. Due to the linearity of the $\mathcal{F}^{-1}$, we obtain:
\begin{equation}
  \mathcal{F}^{-1}[R \cdot e^{j\phi(u,v)}] = R \cdot \mathcal{F}^{-1}[e^{j\phi(u,v)}],  
\end{equation}

which demonstrates that the reconstructed image is a scaled version of the phase-only image, with its structure entirely determined by the phase component.

Next, we apply min-max normalization to the reconstructed image. For an image $\hat{x}^{phase}$, the min-max normalization is defined as:

\begin{equation}
    {x}^{phase} = \frac{\hat{x}^{phase} - \min(\hat{x}^{phase})}{\max(\hat{x}^{phase}) - \min(\hat{x}^{phase})}.
\end{equation}

When we apply the same normalization to the image reconstructed with the constant amplitude $R$, we have:
\begin{align}
    x^{phase'} &= \frac{R \cdot \hat{x}^{phase} - \min(R \cdot \hat{x}^{phase})}
    {\max(R \cdot \hat{x}^{phase}) - \min(R \cdot \hat{x}^{phase})} \\
    &= \frac{R \cdot (\hat{x}^{phase} - \min(\hat{x}^{phase}))}
    {R \cdot (\max(\hat{x}^{phase}) - \min(\hat{x}^{phase}))} \\
    &= x^{phase}.
\end{align}

Thus, we observe that the normalized result $x^{phase'}$ is invariant to the choice of $R$, as long as $R \neq 0$. Therefore, under min-max normalization, the image structure is solely determined by the phase component, while any non-zero constant magnitude $R$ preserves the structural semantics and eliminates domain-specific amplitude variations.

\section{Additional Experiments}\label{sec:appendix B}
\subsection{Domain Prompt Generation Module.}
The Domain Prompt Generation Module involves prompt template design and the impact of K-means clustering on pseudo-label quality, which merit further discussion.

\paragraph{Prompt Templates.}
We use the template ``An image of a $[X]^1\ldots[X]^M$'' for each domain, where the learnable tokens are category-specific and not shared across classes. We analyze category-agnostic compared with category-specific prompts, as well as the impact of the number of learnable tokens $M$.

In fully unsupervised settings, category names are unavailable.
We therefore adopt category-specific phrasing (e.g., “An image of a”) while learning category-conditioned prompt tokens. We further analyze a fully category-agnostic design, where both phrasing and prompt tokens are shared across all categories. In this case, performance degrades: P@50 drops by 3.3\% on DomainNet, and P@1 drops by 9.8\% on Office-Home (65 categories). This result indicates that category-specific prompts are crucial for maintaining discriminability, especially in scenarios with a large number of categories.

\begin{figure}[h]
    \centering
    \includegraphics[height=5cm]{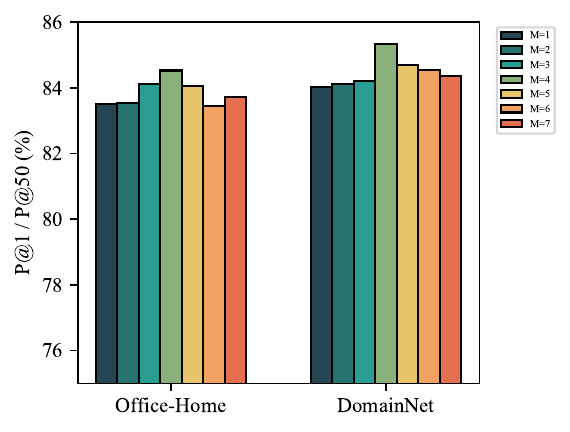}
    \caption{Analysis on the number of learnable text tokens $M$ on two datasets.}
    \label{figure7}
\end{figure}

We analyze the impact of the number of learnable text tokens $M$ on retrieval performance. As illustrated in Figure~\ref{figure7}, the model achieves optimal performance on both datasets when $M=4$. A small value of $M$ may be insufficient to effectively capture rich class-specific semantics and provide adequate supervisory signals. Conversely, setting $M$ too large can result in overfitting of the text prompts and a substantial increase in training time, ultimately hindering generalization.
\paragraph{Pseudo-label Quality.}
Following existing works, we set the number of clusters $K$ equal to the number of classes in the training set, i.e., 65 for Office-Home and 7 for DomainNet. Therefore, the issue of incorrect number of clusters is avoided during training.

To further examine the impact of using an incorrect number of clusters, we conducted additional experiments on DomainNet by setting $K$ to 5 and 9 instead of the ground-truth value of 7. The average P@50 accuracy dropped to 84.59\% and 85.01\%, respectively, compared to 85.33\% under the correct setting. These results indicate that using an inappropriate cluster number can degrade retrieval performance due to inaccurate pseudo-labels and unreliable domain prompts. Nonetheless, the overall performance remains competitive. We also observe that setting $K$ smaller than the actual number of classes tends to result in more severe misclustering, as multiple distinct categories are forced into a single cluster. In contrast, using a larger $K$ may split some categories into multiple clusters, which is generally less detrimental.

To investigate how misclassified points affect the final performance, we evaluate our method under a fully supervised setting, removing the influence of pseudo-label noise. The results on the DomainNet dataset indicate that under the unsupervised setting, where the clustering accuracy is approximately 85\%, the fully supervised counterpart achieves only a modest improvement of 1.64\% in P@50. This confirms that misclassified instances do affect final performance. However, the relatively small performance gap suggests that our unsupervised framework is sufficiently robust to mitigate the negative effects of noisy pseudo-labels. This robustness can be attributed to the following: although some images are incorrectly clustered, the domain prompts are primarily influenced by the majority of correctly clustered samples, resulting in semantically meaningful representations. Consequently, our method achieves competitive retrieval performance without relying on manual annotations.
\subsection{Ablation Study on the impact of the Phase Prior.}

To validate the effectiveness of the phase prior, we conduct an ablation study comparing full-spectrum phase features with high-frequency (HF) and low-frequency (LF) phase variants in the fusion process.
As shown in Table~\ref{table10}, both HF and LF phase priors improve retrieval performance by reducing cross-domain discrepancies while preserving certain semantic cues.
However, full-spectrum phase features consistently achieve superior performance.
This is because the complete phase representation jointly encodes global spatial layout (low-frequency components) and local structural boundaries (high-frequency components), thereby preserving more holistic and domain-invariant structural semantics.
\begin{table*}[h]
  \caption{Average accuracy (\%) for ablation study on phase prior.}
  \label{table10}
  \centering
  \resizebox{0.7\linewidth}{!}{
  \begin{tabular}{lcccccc}
    \toprule
    \multicolumn{1}{c}{\multirow{2}{*}{Method}} & \multicolumn{3}{c}{Office-Home} & \multicolumn{3}{c}{DomainNet} \\
    \cmidrule(lr){2-4} \cmidrule(lr){5-7}
    & $P@1$ & $P@5$ & $P@15$ & $P@50$ & $P@100$ & $P@200$ \\
    \midrule
    Baseline (w/o Prior) & 81.63&80.13&77.72&
82.05&79.94&75.97 \\
    TPSNet (LF Prior) & 83.71&81.06&78.19
&84.25&83.02&80.10\\
    TPSNet (HF Prior) & 84.02&82.99&80.78&84.37&83.22&80.41\\
    \rowcolor{gray!20}\textbf{TPSNet (Phase Prior)} & \textbf{84.53} & \textbf{83.44} & \textbf{81.29} & \textbf{85.33} & \textbf{84.34} & \textbf{81.91} \\
    \bottomrule
  \end{tabular}}
\end{table*}

Furthermore, to investigate the effect of different fusion strategies for integrating the phase prior, we explore four distinct methods to combine phase features with original image features: concatenation, learnable weights, gate-based fusion, and self-attention. As presented in Table~\ref{table11}, the self-attention-based fusion achieves the best performance, highlighting its superior ability to adaptively emphasize informative components and facilitate robust domain-invariant representation learning.
\begin{table*}[t]
  \caption{Ablation study on fusion strategies of the phase prior on the Office-Home dataset (Concat: concatenation, LW: learnable weights, Gate: gate-based fusion, SA: self-attention).}
  \label{table11}
  \centering
  \resizebox{\linewidth}{!}{
    \begin{tabular}{llcccccccccccc}
    \toprule
    \multicolumn{2}{c}{\multirow{2}{*}{Method}} & \multicolumn{3}{c}{$Art \rightarrow Real$} & \multicolumn{3}{c}{$Real \rightarrow Art$} & \multicolumn{3}{c}{$Art \rightarrow Product$} & \multicolumn{3}{c}{$Product \rightarrow Art$} \\ 
    \cmidrule(r){3-5} \cmidrule(r){6-8} \cmidrule(r){9-11} \cmidrule(r){12-14}
    \multicolumn{2}{c}{ }&$P@1$&$P@5$&$P@15$&$P@1$&$P@5$&$P@15$&$P@1$&$P@5$&$P@15$&$P@1$&$P@5$&$P@15$ \\
    \midrule 
    \multirow{4}{*}{ViT-B}
    &Concat&88.50&88.13&87.23&91.30&\textbf{91.26}&\textbf{87.05}&78.78&77.68&76.14&87.02&85.20&79.04 \\
    &LW&82.86&81.61&79.84&86.39&84.18&78.17&73.01&71.16&67.74&80.83&78.87&70.39

 \\
    &Gate&84.14&82.51&80.56&87.65&85.00&79.25&77.96&76.48&	74.73&79.03&79.04&72.09
 \\
    \rowcolor{gray!20}&SA&\textbf{89.53}&\textbf{88.61}&\textbf{87.77}&\textbf{91.74}&90.71&86.81&\textbf{81.17}&\textbf{80.60}&\textbf{79.43}&\textbf{88.80}&\textbf{86.14}&\textbf{81.01} \\
    \midrule 

    \multicolumn{2}{c}{\multirow{2}{*}{Method}} & \multicolumn{3}{c}{$Clipart \rightarrow Real$} & \multicolumn{3}{c}{$Real \rightarrow Clipart$} & \multicolumn{3}{c}{$Product \rightarrow Real$} & \multicolumn{3}{c}{$Real \rightarrow Product$} \\ 
    \cmidrule(r){3-5} \cmidrule(r){6-8} \cmidrule(r){9-11} \cmidrule(r){12-14}
    \multicolumn{2}{c}{ }&$P@1$&$P@5$&$P@15$&$P@1$&$P@5$&$P@15$&$P@1$&$P@5$&$P@15$&$P@1$&$P@5$&$P@15$ \\
    \midrule 
    
    \multirow{4}{*}{ViT-B}

&Concat&73.20&72.41&71.79&88.41&87.17&84.31&91.60&91.77&91.38&	90.57&90.88&90.03

 \\
    &LW&67.40&64.56&62.16&84.00&82.24&77.66&89.93&87.59&85.44&	86.71&85.87&84.87

 \\
    &Gate&68.98&66.72&64.59&85.54&83.59&79.59&90.61&88.77&	86.11&87.19&86.30&84.95

 \\
    \rowcolor{gray!20}&SA&\textbf{73.36}&\textbf{72.73}&\textbf{71.85}&\textbf{89.03}&\textbf{88.01}&\textbf{84.86}&\textbf{93.26}&\textbf{92.82}&\textbf{91.78}&\textbf{92.10}&\textbf{92.05}&\textbf{91.78} \\
    \midrule 
    \multicolumn{2}{c}{\multirow{2}{*}{Method}} & \multicolumn{3}{c}{$Product \rightarrow Clipart$} & \multicolumn{3}{c}{$Clipart \rightarrow Product$} & \multicolumn{3}{c}{$Art \rightarrow Clipart$} & \multicolumn{3}{c}{$Clipart \rightarrow Art$} \\ 
    \cmidrule(r){3-5} \cmidrule(r){6-8} \cmidrule(r){9-11} \cmidrule(r){12-14}
    \multicolumn{2}{c}{ }&$P@1$&$P@5$&$P@15$&$P@1$&$P@5$&$P@15$&$P@1$&$P@5$&$P@15$&$P@1$&$P@5$&$P@15$ \\
    \midrule 

    \multirow{4}{*}{ViT-B}

    &Concat&\textbf{88.98}&\textbf{87.28}&84.46&72.30&\textbf{72.90}&\textbf{72.74}&\textbf{84.71}&\textbf{82.49}&	79.17&69.14&67.44&64.07
 \\
    &LW&84.84&83.40&80.74&66.16&65.67&64.47&78.99&76.41&73.16&	62.47&59.97&55.52

 \\
    &Gate&84.95&83.37&80.16&67.40&66.05&64.93&79.23&75.14&	70.37&62.66&59.58&53.79
 \\
    \rowcolor{gray!20}&SA&88.31&86.74&\textbf{84.49}&\textbf{72.67}&72.30&71.79&84.55&81.94&\textbf{79.36}&\textbf{69.87}&\textbf{68.67}&\textbf{64.55}\\

    \bottomrule
    \end{tabular}}

\end{table*}

\begin{table*}[t]
  \caption{Ablation study on vision-language pretraining architectures on the Office-Home dataset. The table reports results for each of the 12 cross-domain retrieval scenarios individually.}
  \label{table5}
  \centering
  \resizebox{\linewidth}{!}{
    \begin{tabular}{llcccccccccccc}
    \toprule
    \multicolumn{2}{c}{\multirow{2}{*}{Method}} & \multicolumn{3}{c}{$Art \rightarrow Real$} & \multicolumn{3}{c}{$Real \rightarrow Art$} & \multicolumn{3}{c}{$Art \rightarrow Product$} & \multicolumn{3}{c}{$Product \rightarrow Art$} \\ 
    \cmidrule(r){3-5} \cmidrule(r){6-8} \cmidrule(r){9-11} \cmidrule(r){12-14}
    \multicolumn{2}{c}{ }&$P@1$&$P@5$&$P@15$&$P@1$&$P@5$&$P@15$&$P@1$&$P@5$&$P@15$&$P@1$&$P@5$&$P@15$ \\
    \midrule 

    \multirow{4}{*}{ViT-B}
    &TPSNet (BLIP)~\cite{blip}&50.27&49.54&47.89&58.43&53.34&47.45&	43.68&43.79&42.68&55.67&49.70&42.62
 \\
    &TPSNet (SigLIP)~\cite{siglip}&65.14&63.63&61.62&66.01&62.36&56.33&	57.11&53.01&47.67&44.60&42.53&37.48

 \\
    &TPSNet (SigLIP2)~\cite{siglip2}&67.11&66.25&62.21&68.31&66.87&	63.23&62.11&60.12&58.97&47.54&45.33&44.91

 \\
 \rowcolor{gray!20}&TPSNet (CLIP)&\textbf{89.53}&\textbf{88.61}&\textbf{87.77}&\textbf{91.74}&\textbf{90.71}&\textbf{86.81}&\textbf{81.17}&\textbf{80.60}&\textbf{79.43}&\textbf{88.80}&\textbf{86.14}&\textbf{81.01} \\
    \midrule 

    \multicolumn{2}{c}{\multirow{2}{*}{Method}} & \multicolumn{3}{c}{$Clipart \rightarrow Real$} & \multicolumn{3}{c}{$Real \rightarrow Clipart$} & \multicolumn{3}{c}{$Product \rightarrow Real$} & \multicolumn{3}{c}{$Real \rightarrow Product$} \\ 
    \cmidrule(r){3-5} \cmidrule(r){6-8} \cmidrule(r){9-11} \cmidrule(r){12-14}
    \multicolumn{2}{c}{ }&$P@1$&$P@5$&$P@15$&$P@1$&$P@5$&$P@15$&$P@1$&$P@5$&$P@15$&$P@1$&$P@5$&$P@15$ \\
    \midrule 
    
    \multirow{4}{*}{ViT-B}

    &TPSNet (BLIP)~\cite{blip}&26.03&23.89&22.26&50.40&45.69&	38.95&72.70&69.48&65.94&67.04&65.27&63.13
 \\
    &TPSNet (SigLIP)~\cite{siglip}&36.49&35.17&33.50&61.33&57.89&51.34&58.84&54.87&51.82&67.82&64.78&60.38

 \\
    &TPSNet (SigLIP2)~\cite{siglip2}&38.21&37.01&35.03&62.09&	61.77&60.01&60.00&59.04&57.77&70.32&68.43&66.09
 \\
\rowcolor{gray!20} &TPSNet (CLIP)&\textbf{73.36}&\textbf{72.73}&\textbf{71.85}&\textbf{89.03}&\textbf{88.01}&\textbf{84.86}&\textbf{93.26}&\textbf{92.82}&\textbf{91.78}&\textbf{92.10}&\textbf{92.05}&\textbf{91.78} \\
    \midrule 

    \multicolumn{2}{c}{\multirow{2}{*}{Method}} & \multicolumn{3}{c}{$Product \rightarrow Clipart$} & \multicolumn{3}{c}{$Clipart \rightarrow Product$} & \multicolumn{3}{c}{$Art \rightarrow Clipart$} & \multicolumn{3}{c}{$Clipart \rightarrow Art$} \\ 
    \cmidrule(r){3-5} \cmidrule(r){6-8} \cmidrule(r){9-11} \cmidrule(r){12-14}
    \multicolumn{2}{c}{ }&$P@1$&$P@5$&$P@15$&$P@1$&$P@5$&$P@15$&$P@1$&$P@5$&$P@15$&$P@1$&$P@5$&$P@15$ \\
    \midrule 
    
    \multirow{4}{*}{ViT-B}

    &TPSNet (BLIP)~\cite{blip}&56.14&49.97&42.35&24.56&	24.33&24.09&35.15&31.91&28.22&18.99&17.23&14.92 \\
    &TPSNet (SigLIP)~\cite{siglip}&45.57&43.77&39.08&32.97&32.20&29.92&55.79&52.01&45.87&30.65&28.20&25.31\\
    &TPSNet (SigLIP2)~\cite{siglip2}&49.32&48.88&47.61&	33.21&32.90&30.05&57.09&54.87&53.21&32.64&30.40&28.65
 \\
 \rowcolor{gray!20}&TPSNet (CLIP)&\textbf{88.31}&\textbf{86.74}&\textbf{84.49}&\textbf{72.67}&\textbf{72.30}&\textbf{71.79}&\textbf{84.55}&\textbf{81.94}&\textbf{79.36}&\textbf{69.87}&\textbf{68.67}&\textbf{64.55} \\
    
    \bottomrule
    \end{tabular}}

\end{table*}
\begin{table*}[t]
  \caption{Ablation study on vision-language pretraining architectures on the Domainnet dataset. The table reports results for each of the 12 cross-domain retrieval scenarios individually.}
  \label{table6}
  \centering
  \resizebox{\linewidth}{!}{
    \begin{tabular}{llcccccccccccc}
    \toprule
    \multicolumn{2}{c}{\multirow{2}{*}{Method}} & \multicolumn{3}{c}{$Clipart \rightarrow Sketch$} & \multicolumn{3}{c}{$Sketch \rightarrow Clipart$} & \multicolumn{3}{c}{$Infograph \rightarrow Real$} & \multicolumn{3}{c}{$Real \rightarrow Infograph$} \\ 
    \cmidrule(r){3-5} \cmidrule(r){6-8} \cmidrule(r){9-11} \cmidrule(r){12-14}
    \multicolumn{2}{c}{ }&$P@50$&$P@100$&$P@200$&$P@50$&$P@100$&$P@200$&$P@50$&$P@100$&$P@200$&$P@50$&$P@100$&$P@200$ \\
    \midrule 
    
    \multirow{4}{*}{ViT-B}
    &TPSNet (BLIP)~\cite{blip}&81.48&79.61&72.53&77.04&75.26&68.83&	58.33&58.21&57.84&87.03&78.33&64.53

 \\
    &TPSNet (SigLIP)~\cite{siglip}&84.37&79.48&73.07&81.02&76.99&70.93&	63.19&62.90&62.43&73.55&64.03&50.73

\\
    &TPSNet (SigLIP2)~\cite{siglip2}&89.42&86.18&81.06&88.24&85.88&	80.63&73.02&72.85&72.67&80.28&78.06&66.05

 \\
 \rowcolor{gray!20}&TPSNet (CLIP)&\textbf{97.92}&\textbf{97.84}&\textbf{97.73}&\textbf{98.27}&\textbf{98.20}&\textbf{98.06}&\textbf{78.72}&\textbf{78.55}&\textbf{78.31}&\textbf{93.01}&\textbf{93.04}&\textbf{85.71} \\
    \midrule 

    \multicolumn{2}{c}{\multirow{2}{*}{Method}} & \multicolumn{3}{c}{$Infograph \rightarrow Sketch$} & \multicolumn{3}{c}{$Sketch \rightarrow Infograph$} & \multicolumn{3}{c}{$Painting \rightarrow Clipart$} & \multicolumn{3}{c}{$Clipart \rightarrow Painting$} \\ 
    \cmidrule(r){3-5} \cmidrule(r){6-8} \cmidrule(r){9-11} \cmidrule(r){12-14}
    \multicolumn{2}{c}{ }&$P@50$&$P@100$&$P@200$&$P@50$&$P@100$&$P@200$&$P@50$&$P@100$&$P@200$&$P@50$&$P@100$&$P@200$ \\
    
    \midrule 
    \multirow{4}{*}{ViT-B}

    &TPSNet (BLIP)~\cite{blip}&50.91&48.61&43.12&59.20&52.93&	44.21&86.19&84.71&76.74&81.09&79.39&75.76 \\
    &TPSNet (SigLIP)~\cite{siglip}&59.46&56.58&53.44&70.11&65.79&	62.02&88.79&81.97&79.91&86.50&84.51&81.63\\
    &TPSNet (SigLIP2)~\cite{siglip2}&64.64&61.40&55.93&73.45&	70.22&64.87&94.83&88.82&88.25&89.42&88.75&87.81
 \\
 \rowcolor{gray!20}&TPSNet (CLIP)&\textbf{75.87}&\textbf{75.61}&\textbf{75.03}&\textbf{95.21}&\textbf{92.04}&\textbf{84.78}&\textbf{98.49}&\textbf{98.45}&\textbf{98.26}&\textbf{97.84}&\textbf{97.73}&\textbf{96.64} \\
    \midrule 

    \multicolumn{2}{c}{\multirow{2}{*}{Method}} & \multicolumn{3}{c}{$Painting \rightarrow Quickdraw$} & \multicolumn{3}{c}{$Quickdraw \rightarrow Painting$} & \multicolumn{3}{c}{$Quickdraw \rightarrow Real$} & \multicolumn{3}{c}{$Real \rightarrow Quickdraw$} \\ 
    \cmidrule(r){3-5} \cmidrule(r){6-8} \cmidrule(r){9-11} \cmidrule(r){12-14}
    \multicolumn{2}{c}{ }&$P@50$&$P@100$&$P@200$&$P@50$&$P@100$&$P@200$&$P@50$&$P@100$&$P@200$&$P@50$&$P@100$&$P@200$ \\
    
    \midrule 
    \multirow{4}{*}{ViT-B}

    &TPSNet (BLIP)~\cite{blip}&32.06&30.40&28.18&23.75&	22.97&22.07&22.34&23.08&23.36&34.92&32.75&30.16 \\
    &TPSNet (SigLIP)~\cite{siglip}&37.21&36.46&35.51&18.47&	17.49&16.69&19.60&18.60&17.71&26.48&25.84&24.67\\
    &TPSNet (SigLIP2)~\cite{siglip2}&37.04&36.22&35.04&	17.22&17.23&17.36&17.66&17.21&17.45&32.45&30.56&29.04
 \\
 \rowcolor{gray!20}&TPSNet (CLIP)&\textbf{81.67}&\textbf{77.04}&\textbf{71.34}&\textbf{65.12}&\textbf{64.51}&\textbf{63.31}&\textbf{62.50}&\textbf{62.21}&\textbf{61.56}&\textbf{79.34}&\textbf{76.89}&\textbf{72.14} \\
    \bottomrule
    \end{tabular}}
\end{table*}
\begin{table*}[t]
  \centering
  \caption{Average accuracy (\%) compared with universal retrieval models on the Office-Home and DomainNet Datasets.}
  \label{table7}
  \resizebox{0.7\linewidth}{!}{   
    \begin{tabular}{lcccccc}
      \toprule
      \multicolumn{1}{c}{\multirow{2}{*}{Method}} & \multicolumn{3}{c}{Office-Home} & \multicolumn{3}{c}{DomainNet} \\
      \cmidrule(lr){2-4} \cmidrule(lr){5-7}
       & $P@1$ & $P@5$ & $P@15$ & $P@50$ & $P@100$ & $P@200$ \\
      \midrule
      TPSNet (BLIP)~\cite{blip} & 46.59 & 43.68 & 40.04 & 57.86 & 55.52 & 50.61 \\
      TPSNet (SigLIP)~\cite{siglip} & 51.86 & 49.20 & 45.03 & 59.06 & 55.89 & 52.40 \\
      TPSNet (SigLIP2)~\cite{siglip2} & 54.00 & 52.66 & 50.65 & 63.14 & 61.12 & 58.01 \\
      \rowcolor{gray!20}\textbf{TPSNet (CLIP)} & \textbf{84.53} & \textbf{83.44} & \textbf{81.29} & \textbf{85.33} & \textbf{84.34} & \textbf{81.91} \\
      \bottomrule
    \end{tabular}
  }
\end{table*}

\subsection{Ablation Study on the Impact of Vision-Language Pretraining Architectures.}

We further investigate the impact of different vision-language pretraining architectures on the performance of TPSNet. Specifically, we compare CLIP~\cite{clip}, BLIP~\cite{blip}, SigLIP~\cite{siglip} and SigLIP2~\cite{siglip2} as backbone models. The experimental results on the OfficeHome and DomainNet datasets are presented in Table~\ref{table5} and Table~\ref{table6}, with the average performance summarized in Table~\ref{table7}. Empirical results reveal that TPSNet achieves the best performance with CLIP as the backbone. In contrast, its performance noticeably degrades when using BLIP. We attribute this to BLIP’s multi-task pretraining objective, which incorporates both image-text matching and image-conditioned text generation. While such generative objectives enhance performance in tasks like image captioning and visual question answering, they may limit the model’s ability to learn fine-grained, discriminative representations that are critical for cross-domain image retrieval tasks.

Similarly, SigLIP also exhibits inferior performance compared to CLIP, despite adopting a contrastive learning framework. We attribute this discrepancy to differences in their loss function formulations. Specifically, SigLIP employs a sigmoid-based contrastive loss that treats each image-text pair independently, whereas CLIP utilizes a softmax-based loss that jointly considers all positive and negative pairs within a batch. This global contrastive objective in CLIP fosters stronger alignment and discrimination between modalities, which in turn enhances the model’s capacity to learn semantically consistent representations across domains. SigLIP2~\cite{siglip2}, an improved variant of SigLIP, mitigates some of these limitations by introducing architectural enhancements and refined training strategies, leading to consistently better performance than SigLIP across two datasets. However, its performance remains inferior to CLIP, primarily because the sigmoid-based contrastive loss fails to capture the global batch-level interactions inherent in CLIP’s softmax-based objective.

\subsection{Ablation Study on the Impact of Image Encoder Initialization.}\label{sec:ablation_encoder}

\begin{table*}[h]
  \caption{Ablation study on the impact of image encoder initialization on the Office-Home dataset. The table reports results for each of the 12 cross-domain retrieval scenarios individually.}
  \label{table8}
  \centering
  \resizebox{\linewidth}{!}{
    \begin{tabular}{llcccccccccccc}
    \toprule
    \multicolumn{2}{c}{\multirow{2}{*}{Method}} & \multicolumn{3}{c}{$Art \rightarrow Real$} & \multicolumn{3}{c}{$Real \rightarrow Art$} & \multicolumn{3}{c}{$Art \rightarrow Product$} & \multicolumn{3}{c}{$Product \rightarrow Art$} \\ 
    \cmidrule(r){3-5} \cmidrule(r){6-8} \cmidrule(r){9-11} \cmidrule(r){12-14}
    \multicolumn{2}{c}{ }&$P@1$&$P@5$&$P@15$&$P@1$&$P@5$&$P@15$&$P@1$&$P@5$&$P@15$&$P@1$&$P@5$&$P@15$ \\
    \midrule 
    \multirow{3}{*}{ResNet-50}
    &ShieldIR (SOTA)~\cite{shieldir}&51.46 &50.20& 49.27& 57.33& 53.50 &49.26 &44.75& 46.16 &45.33 &56.31& 52.71& 47.39 \\
    &TPSNet (MoCov2)&60.94&60.06&59.04&65.89& 62.10& 57.50& 53.07& 50.67& 48.89& 60.24& 56.92& 49.49
 \\
    \rowcolor{gray!20}&TPSNet&\textbf{69.06}&\textbf{67.22}&\textbf{65.02}&\textbf{74.00}&\textbf{69.72}&\textbf{62.73}&\textbf{59.41}&\textbf{57.21}&\textbf{53.50}&\textbf{63.82}&\textbf{59.98}&\textbf{52.31} \\
    \midrule 
    \multirow{3}{*}{ViT-B}
    &SA-MoE (SOTA)~\cite{samoe}&71.12&68.93&66.10&73.86&68.85&60.37&64.69&62.39&58.63&66.57&62.82&55.12 \\
    &TPSNet (DINO)& 73.22& 71.54& 70.50& 79.27& 76.15& 69.79& 64.98& 64.23& 62.85& 74.32& 69.94& 63.11
 \\
    \rowcolor{gray!20}&TPSNet&\textbf{89.53}&\textbf{88.61}&\textbf{87.77}&\textbf{91.74}&\textbf{90.71}&\textbf{86.81}&\textbf{81.17}&\textbf{80.60}&\textbf{79.43}&\textbf{88.80}&\textbf{86.14}&\textbf{81.01} \\
    \midrule 

    \multicolumn{2}{c}{\multirow{2}{*}{Method}} & \multicolumn{3}{c}{$Clipart \rightarrow Real$} & \multicolumn{3}{c}{$Real \rightarrow Clipart$} & \multicolumn{3}{c}{$Product \rightarrow Real$} & \multicolumn{3}{c}{$Real \rightarrow Product$} \\ 
    \cmidrule(r){3-5} \cmidrule(r){6-8} \cmidrule(r){9-11} \cmidrule(r){12-14}
    \multicolumn{2}{c}{ }&$P@1$&$P@5$&$P@15$&$P@1$&$P@5$&$P@15$&$P@1$&$P@5$&$P@15$&$P@1$&$P@5$&$P@15$ \\
    \midrule 
    \multirow{3}{*}{ResNet-50}
    &ShieldIR (SOTA)~\cite{shieldir}&43.36& 42.58& 41.55& 52.05& 52.38 &50.81 &70.37 &68.33 &66.08 &61.46& 62.08 &61.85\\
    &TPSNet (MoCov2)& 46.30& 45.45& 44.66& 59.03& 57.49& 54.19& 76.57& 73.56& 71.22& 71.15& 69.66& 68.25
 \\
    \rowcolor{gray!20}&TPSNet&\textbf{51.68}&\textbf{49.91}&\textbf{47.56}&\textbf{63.53}&\textbf{61.36}&\textbf{57.40}&\textbf{78.85}&\textbf{75.87}&\textbf{72.48}&\textbf{74.39}&\textbf{73.00}&\textbf{70.94} \\
    \midrule 
    \multirow{3}{*}{ViT-B}

    &SA-MoE (SOTA)~\cite{samoe}&52.99&50.00&47.33&70.41&65.78&60.06&78.91&74.96&70.40&76.52&74.54&71.18 \\
    &TPSNet (DINO)& 54.41& 52.77& 51.86& 73.35& 70.38& 66.94& 82.18& 80.61& 78.45& 79.94& 78.32& 76.59
 \\
    \rowcolor{gray!20}&TPSNet&\textbf{73.36}&\textbf{72.73}&\textbf{71.85}&\textbf{89.03}&\textbf{88.01}&\textbf{84.86}&\textbf{93.26}&\textbf{92.82}&\textbf{91.78}&\textbf{92.10}&\textbf{92.05}&\textbf{91.78} \\
    \midrule 

    \multicolumn{2}{c}{\multirow{2}{*}{Method}} & \multicolumn{3}{c}{$Product \rightarrow Clipart$} & \multicolumn{3}{c}{$Clipart \rightarrow Product$} & \multicolumn{3}{c}{$Art \rightarrow Clipart$} & \multicolumn{3}{c}{$Clipart \rightarrow Art$} \\ 
    \cmidrule(r){3-5} \cmidrule(r){6-8} \cmidrule(r){9-11} \cmidrule(r){12-14}
    \multicolumn{2}{c}{ }&$P@1$&$P@5$&$P@15$&$P@1$&$P@5$&$P@15$&$P@1$&$P@5$&$P@15$&$P@1$&$P@5$&$P@15$ \\
    \midrule 
    \multirow{3}{*}{ResNet-50}
    &ShieldIR (SOTA)~\cite{shieldir}&52.10& 52.43 &50.23 &40.84 &39.87 &39.45 &41.24 &39.05& 37.44& 32.21& 30.13& 28.39\\
    &TPSNet (MoCov2)& 59.92&\textbf{57.74}&\textbf{56.20}& 42.54& 42.68&\textbf{42.37}& 48.53& 45.74& 42.67& 36.38& 33.58& 31.20 \\
    \rowcolor{gray!20}&TPSNet&\textbf{60.71}& 57.49& 53.26&\textbf{46.05}&\textbf{43.77}& 42.22&\textbf{56.98}&\textbf{54.07}&\textbf{50.29}&\textbf{43.14}&\textbf{39.66}&\textbf{35.58} \\
    \midrule 
    \multirow{3}{*}{ViT-B}

    &SA-MoE (SOTA)~\cite{samoe}&65.28&61.09&56.01&47.47&45.81&43.36&61.80&57.82&53.23& 46.90& 44.39& 39.97 \\
    &TPSNet (DINO)& 72.38& 68.05& 64.58& 49.90& 49.23& 47.57& 63.12& 60.66& 56.65&44.42&	41.87&38.56\\
    \rowcolor{gray!20}&TPSNet&\textbf{88.31}&\textbf{86.74}&\textbf{84.49}&\textbf{72.67}&\textbf{72.30}&\textbf{71.79}&\textbf{84.55}&\textbf{81.94}&\textbf{79.36}&\textbf{69.87}&\textbf{68.67}&\textbf{64.55} \\
    
    \bottomrule
    \end{tabular}}

\end{table*}
\begin{table*}[h]
  \caption{Ablation study on the impact of image encoder initialization on the Domainnet dataset. The table reports results for each of the 12 cross-domain retrieval scenarios individually.}
  \label{table9}
  \centering
  \resizebox{\linewidth}{!}{
    \begin{tabular}{llcccccccccccc}
    \toprule
    \multicolumn{2}{c}{\multirow{2}{*}{Method}} & \multicolumn{3}{c}{$Clipart \rightarrow Sketch$} & \multicolumn{3}{c}{$Sketch \rightarrow Clipart$} & \multicolumn{3}{c}{$Infograph \rightarrow Real$} & \multicolumn{3}{c}{$Real \rightarrow Infograph$} \\ 
    \cmidrule(r){3-5} \cmidrule(r){6-8} \cmidrule(r){9-11} \cmidrule(r){12-14}
    \multicolumn{2}{c}{ }&$P@50$&$P@100$&$P@200$&$P@50$&$P@100$&$P@200$&$P@50$&$P@100$&$P@200$&$P@50$&$P@100$&$P@200$ \\
    \midrule 
    \multirow{3}{*}{ResNet-50}
    &ShieldIR (SOTA)~\cite{shieldir}&87.16 &87.12& 86.74 &89.95& 89.54 &88.56 &54.41 &54.93 &55.34 &82.26& 75.35 &64.15\\
    &TPSNet (MoCov2)&\textbf{95.64}&\textbf{95.85}&\textbf{95.93}&\textbf{96.13}&\textbf{96.08}&\textbf{96.03}&47.28&47.46&47.40& 83.82& 76.42& 61.52\\
    \rowcolor{gray!20}&TPSNet& 95.43& 95.31& 94.63& 95.50& 95.31& 94.26&\textbf{70.04}&\textbf{69.55}&\textbf{68.74}&\textbf{86.07}&\textbf{81.91}&\textbf{71.99}\\
    \midrule 
    \multirow{3}{*}{ViT-B}
    &SA-MoE (SOTA)~\cite{samoe}&83.97& 82.08& 78.40& 88.11& 86.07& 81.21& 57.29 &57.55& 57.67& 87.37& 80.08 &66.24 \\
    &TPSNet (DINO)& 96.18& 96.27& 96.19& 97.18& 97.13& 96.82& 61.54& 61.47& 61.35& 90.21& 83.78& 69.13\\
    \rowcolor{gray!20}&TPSNet&\textbf{97.92}&\textbf{97.84}&\textbf{97.73}&\textbf{98.27}&\textbf{98.20}&\textbf{98.06}&\textbf{78.72}&\textbf{78.55}&\textbf{78.31}&\textbf{93.01}&\textbf{93.04}&\textbf{85.71} \\
    \midrule 

    \multicolumn{2}{c}{\multirow{2}{*}{Method}} & \multicolumn{3}{c}{$Infograph \rightarrow Sketch$} & \multicolumn{3}{c}{$Sketch \rightarrow Infograph$} & \multicolumn{3}{c}{$Painting \rightarrow Clipart$} & \multicolumn{3}{c}{$Clipart \rightarrow Painting$} \\ 
    \cmidrule(r){3-5} \cmidrule(r){6-8} \cmidrule(r){9-11} \cmidrule(r){12-14}
    \multicolumn{2}{c}{ }&$P@50$&$P@100$&$P@200$&$P@50$&$P@100$&$P@200$&$P@50$&$P@100$&$P@200$&$P@50$&$P@100$&$P@200$ \\
    \midrule 
    \multirow{3}{*}{ResNet-50}
    &ShieldIR (SOTA)~\cite{shieldir}&47.39 &47.10 &46.34 &72.29& 67.68 &55.00 &93.96 &93.47 &90.99 &85.32& 85.34 &84.30\\
    &TPSNet (MoCov2)& 46.86& 46.66& 45.82& 82.04& 73.52& 58.22&\textbf{95.50}&\textbf{95.00}&\textbf{93.95}& 92.83& 92.55&\textbf{91.70}\\
    \rowcolor{gray!20}&TPSNet&\textbf{67.59}&\textbf{67.00}&\textbf{65.96}&\textbf{90.20}&\textbf{86.27}&\textbf{74.90}& 94.07& 93.27& 90.61&\textbf{94.56}&\textbf{93.99}& 91.08\\
    \midrule 
    \multirow{3}{*}{ViT-B}

    &SA-MoE (SOTA)~\cite{samoe}&53.76& 53.07& 50.37& 79.56& 72.29& 58.62& 95.00& 94.59& 93.23& 90.60& 90.42& 89.25 \\
    &TPSNet (DINO)& 60.03& 59.92& 59.44& 91.94& 85.83& 69.64& 97.89& 97.68& 96.77& 97.11& 96.89& 95.29\\
    \rowcolor{gray!20}&TPSNet&\textbf{75.87}&\textbf{75.61}&\textbf{75.03}&\textbf{95.21}&\textbf{92.04}&\textbf{84.78}&\textbf{98.49}&\textbf{98.45}&\textbf{98.26}&\textbf{97.84}&\textbf{97.73}&\textbf{96.64} \\
    \midrule 

    \multicolumn{2}{c}{\multirow{2}{*}{Method}} & \multicolumn{3}{c}{$Painting \rightarrow Quickdraw$} & \multicolumn{3}{c}{$Quickdraw \rightarrow Painting$} & \multicolumn{3}{c}{$Quickdraw \rightarrow Real$} & \multicolumn{3}{c}{$Real \rightarrow Quickdraw$} \\ 
    \cmidrule(r){3-5} \cmidrule(r){6-8} \cmidrule(r){9-11} \cmidrule(r){12-14}
    \multicolumn{2}{c}{ }&$P@50$&$P@100$&$P@200$&$P@50$&$P@100$&$P@200$&$P@50$&$P@100$&$P@200$&$P@50$&$P@100$&$P@200$ \\
    \midrule 
    \multirow{3}{*}{ResNet-50}
    &ShieldIR (SOTA)~\cite{shieldir}&65.33 &61.25& 57.98& 58.07 &57.42& \textbf{55.11}& 67.19 &67.36 &67.53& \textbf{75.70}& 73.60 &70.63\\
    &TPSNet (MoCov2)& 64.56& 63.20& 61.86&46.74&48.92&49.88& 65.81& 66.40& 66.97& 71.26& 71.54& 70.75 \\
    \rowcolor{gray!20}&TPSNet&\textbf{67.21}&\textbf{65.38}&\textbf{61.88}&\textbf{59.27}&\textbf{57.93}&54.30&\textbf{69.34}&\textbf{68.88}&\textbf{68.17}&75.06&\textbf{74.70}&\textbf{72.96} \\
    \midrule 
    \multirow{3}{*}{ViT-B}

    &SA-MoE (SOTA)~\cite{samoe}& 72.54&  71.16& 69.22&61.32& 61.21& 61.03& 55.78& 55.59& 55.49& 59.02& 57.55& 55.81 \\
    &TPSNet (DINO)&69.00&67.33&64.19&62.70&60.42&56.22&\textbf{68.98}&\textbf{67.80}&\textbf{65.70}& 78.93&\textbf{77.04}&\textbf{75.04}\\
    \rowcolor{gray!20}&TPSNet&\textbf{81.67}&\textbf{77.04}&\textbf{71.34}&\textbf{65.12}&\textbf{64.51}&\textbf{63.31}& 62.50& 62.21& 61.56&\textbf{79.34}& 76.89& 72.14 \\
    \bottomrule
    \end{tabular}}
\end{table*}

To evaluate the effectiveness of TPSNet without attributing the observed improvements to the CLIP image encoder, we replace the CLIP pre-trained image encoder with MoCov2~\cite{moco} and DINO~\cite{DINO}, which have been widely adopted in previous works~\cite{ucdir,protoot,samoe}. This setup ensures that any performance gains are solely due to the effectiveness of TPSNet itself. The detailed experimental results are provided in Tables~\ref{table8} and~\ref{table9}.

As demonstrated in Table~\ref{table8} and Table~\ref{table9}, even when MoCov2 and DINO are employed as initialization parameters for the image encoder, TPSNet consistently outperforms current state-of-the-art methods on both datasets. Furthermore, when the image encoder is initialized with parameters pre-trained by CLIP, TPSNet's performance is further enhanced.
\subsection{Dynamic Prompt Generation vs. Our Domain Prompts}

For prompt generation, we also considered alternative learnable prompt-generation strategies. Regarding the use of pre-trained vision–language models to automatically refine prompts, we experimented with employing QwenVL to generate class-relevant prompts directly from images. However, we found that such models often produce semantically inconsistent textual outputs for instances within the same class. For example, images labeled as “bird” may yield prompts such as “a parrot” or “a hummingbird”, leading to semantic fragmentation that severely undermines retrieval consistency. Moreover, these methods typically depend on manually crafted templates (e.g., “What is the specific object in this image?”) to guide prompt generation, which introduces additional engineering overhead. We also observed that each inference takes approximately 0.91 seconds and consumes more than 4 GB of GPU memory per image, rendering this approach computationally expensive and impractical for large-scale unsupervised datasets.

Therefore, in our setting, we adopt a learnable prompt format of “An image of a $[X]^1[X]^2\ldots[X]^M$”, which does not rely on class labels. For each class, we generate a more descriptive yet consistent prompt expression that avoids the semantic fragmentation issue of dynamically generated prompts while remaining robust across domains. We believe this design provides a practical and effective balance between performance and training cost.

\subsection{Limitations of the LMM-based Universal Retrieval Models for UCDIR.}
In addition, we compare TPSNet with the current state-of-the-art large multimodal models (LMMs) for retrieval performance, including InternVL3.5~\cite{internvl}, SAIL-VL2~\cite{sail}, MM-Embed~\cite{mmembed}, Qwen2.5VL~\cite{qwen2.5}, and LamRA~\cite{lamra}. The average retrieval performance is summarized in Table~\ref{table12}. Since InternVL3.5 and SAIL-VL2 are primarily designed for question-answering tasks, we use only their pre-trained vision encoders. All other methods utilize the complete output vectors of LMMs as feature representations. Experimental results demonstrate that, although LMM-based universal retrieval models exhibit strong zero-shot generalization ability, there remains a noticeable performance gap compared to TPSNet, particularly in scenarios involving significant domain discrepancies. Specifically, their performance deteriorates significantly in domains with substantial background noise and abstract visual patterns, such as Infograph and Quickdraw. This degradation can be attributed to the inherent bias of foundation models towards natural-image domains prevalent in their pretraining datasets, which limits their ability to generalize to highly stylized domains.

\begin{table*}[t!]
  \caption{Average accuracy (\%) compared with the LMM-based universal retrieval models on the Office-Home and DomainNet Datasets.}
  \label{table12}
  \centering
  \resizebox{0.8\linewidth}{!}{
  \begin{tabular}{lccccccc}
    \toprule
    \multicolumn{1}{c}{\multirow{2}{*}{Method}} & \multicolumn{3}{c}{Office-Home} & \multicolumn{3}{c}{DomainNet} \\
    \cmidrule(lr){2-4} \cmidrule(lr){5-7}
    &$P@1$ & $P@5$ & $P@15$ & $P@50$ & $P@100$ & $P@200$ \\
    \midrule
      InternVL3.5 - 6B~\cite{internvl} &66.08&60.88&53.81
&58.71&54.82&47.11\\
    SAIL-VL2 - 0.6B~\cite{sail} & 73.08&67.47&58.76
&58.29&53.38&46.16
 \\
    MM-Embed - 7B~\cite{mmembed} & 79.38&76.20&71.01&78.05&75.18&70.27
 \\
 Qwen2.5VL - 3B~\cite{qwen2.5} & 77.84&75.01&69.40
&78.00&77.25&75.81

 \\
  Qwen2.5VL - 7B~\cite{qwen2.5} & 82.56&80.53&76.12
&81.45&81.47&80.44

 \\

    LamRA - 7B~\cite{lamra}& 84.20&82.60&79.75&80.77&78.39&74.80

 \\
    \rowcolor{gray!20}\textbf{TPSNet - 0.086B} & \textbf{84.53} & \textbf{83.44} & \textbf{81.29} & \textbf{85.33} & \textbf{84.34} & \textbf{81.91} \\
    \bottomrule
  \end{tabular}}
\end{table*}

Therefore, despite their strong generalization capacity, universal retrieval models still fall short in effectively addressing the specific challenges of UCDIR. Moreover, these models typically require a substantial number of parameters, whereas our TPSNet, based on ViT-B, achieves superior performance with a significantly reduced parameter count. This demonstrates the necessity and practical value of developing targeted methods tailored for UCDIR, especially in scenarios with severe domain discrepancies.

\begin{figure}[t]
  \centering
  \includegraphics[height=4cm]{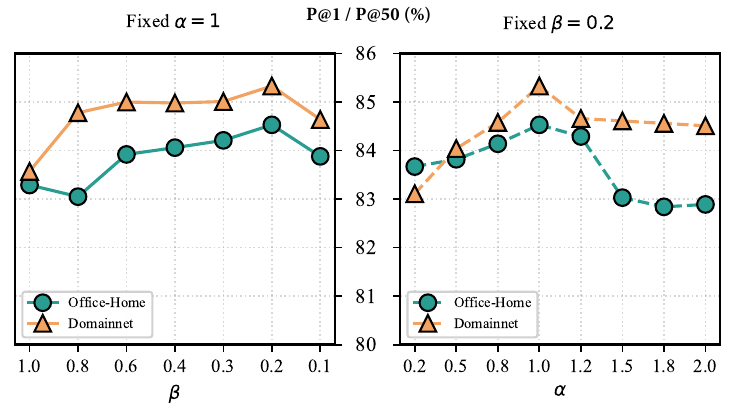}
  \vspace{-0.5em}
  \caption{Analysis on coefficients $\alpha$ and $\beta$ on two datasets.}
  \label{figure8a}
\end{figure}

\begin{figure}[t]
  \centering
  \includegraphics[height=4cm]{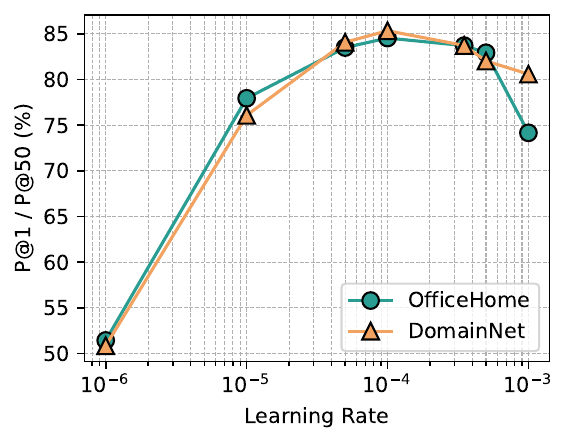}
  \vspace{-0.8em}
  \caption{Analysis on learning rate.}
  \label{figure8b}
  
\end{figure}

\subsection{Analysis on Coefficients $\alpha$ and $\beta$.}
To investigate the impact of the coefficients $\alpha$ and $\beta$ in Eq.\eqref{loss}, we conduct hyperparameter analysis experiments on Office-Home and Domainnet datasets, as shown in Figure~\ref{figure8a}. The model achieves the best performance when $\alpha=1$ and $\beta=0.2$.

\subsection{Analysis on Learning Rate.}
We conduct a learning rate sensitivity analysis on both the Office-Home and DomainNet datasets to examine its impact on model performance. As shown in Figure~\ref{figure8b}, the best performance is achieved when the learning rate is set to $10^{-4}$.

\subsection{More Visualization Results}
\begin{figure*}[t]
    \centering
    \includegraphics[width=\linewidth]{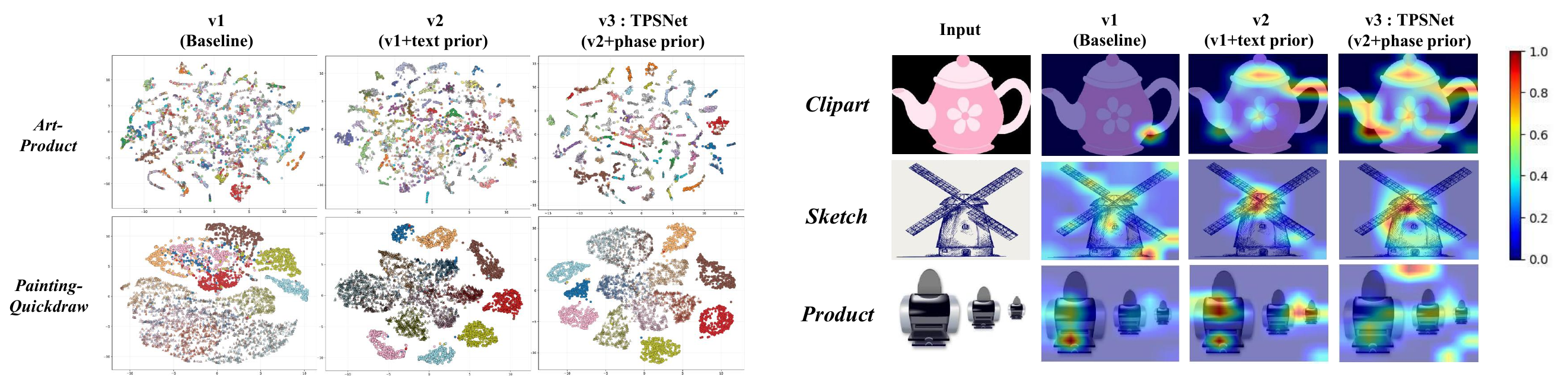}
    \caption{Additional t-SNE and Grad-CAM visualizations of last-layer features for the baseline model (v1), baseline with text prior (v2), and TPSNet (v3) across two datasets.}
    \label{figure10}
    \vspace{-0.5em}
\end{figure*}

Here, we present additional t-SNE and Grad-CAM visualizations across more scenarios and samples in Figure~\ref{figure10}, demonstrating the explainability of our TPSNet method on multiple cross-domain tasks and datasets.

\subsection{Computational Cost Analysis}\label{sec:cost}
We evaluate the additional computational overhead introduced by the phase prior extraction and text prior cross-modal attention mechanisms. To quantify this overhead, we conduct detailed ablation studies on four key metrics: parameter count, FLOPs, inference time, and memory usage. The results are summarized in Table~\ref{table13}.

\begin{table*}[h]
  \centering
  \caption{The additional computational overhead introduced by the phase spectrum extraction and cross-modal attention mechanisms.}
  \label{table13}
  \setlength{\tabcolsep}{3pt}
  \resizebox{0.7\linewidth}{!}{
  \begin{tabular}{ccccccc}
    \toprule
    DPG & TPFE & PPFE & Params\_all & FLOPs & Inference Time & Memory Usage \\ 
    \midrule 
    \XSolidBrush & \XSolidBrush & \XSolidBrush & 114.518M & 45.081G & 0.031ms & 1067.23MB \\ 
    \XSolidBrush   & \XSolidBrush & \Checkmark   & 115.549M & 47.042G & 0.037ms & 1104.28MB \\ 
    \Checkmark   & \Checkmark   & \XSolidBrush & 115.569M & 45.088G & 0.095ms & 1085.37MB \\ 
    \rowcolor{gray!20}
    \Checkmark   & \Checkmark   & \Checkmark   & 116.600M & 47.049G & 0.104ms & 1119.16MB \\
    \bottomrule 
  \end{tabular}
  }
\end{table*}

As shown in our ablation study, the phase branch—though built with a lightweight convolutional encoder—introduces an additional 1.03M parameters and 1.96G FLOPs, primarily due to processing an extra phase image in parallel with the RGB input. Despite this, the increase in inference time is negligible (+0.006ms), and the memory overhead remains modest (+37MB), demonstrating the efficiency of the design. In contrast, text prior cross-modal attention mechanism adds only 0.02M parameters and 0.01G FLOPs, as it reuses existing features and performs lightweight token-level cross-modal interactions. However, it results in a more noticeable increase in inference time (+0.064ms), attributed to the sequential attention operations—including Q-K-V projections, matrix multiplications, and softmax computations—which are computationally intensive and less parallelizable than convolutional operations. Overall, the full model introduces just 2.08M additional parameters (1.8\%) and 2.0G more FLOPs (4.4\%) compared to the baseline. Inference time rises from 0.031ms to 0.104ms, and memory usage increases moderately from 1067MB to 1119MB. These results demonstrate that our proposed method maintains high efficiency and is well-suited for scalable deployment on larger datasets or in real-time applications.

\section{Limitations}

Although TPSNet achieves significant improvements on the UCDIR task, it still has certain limitations. In DPGM, the generation of domain prompts relies on a manually designed prompt in the form of ``An image of a $[X]^1[X]^2\ldots[X]^M$.'' While this provides a simple and interpretable way to encode image semantics, it may limit the expressiveness and flexibility of the prompts, especially in complex or diverse domains. In future work, we plan to explore more dynamic or learnable prompt generation strategies to enhance generalization across broader domain distributions. 

\end{document}